\documentclass[10pt,twocolumn,letterpaper]{article}

\usepackage{iccv}
\usepackage{times}
\usepackage{epsfig}
\usepackage{graphicx}
\usepackage{amsmath}
\usepackage{amssymb}

\usepackage{subfigure}
\usepackage{epsfig}
\usepackage{multirow}
\usepackage{wrapfig}
\usepackage{mathtools,cuted}
\usepackage{helvet}
\usepackage{courier}
\usepackage{amsmath}
\usepackage{bm}
\usepackage{dsfont}
\usepackage{paralist}

\input{taesup.sty}

\usepackage[pagebackref=true,breaklinks=true,letterpaper=true,colorlinks,bookmarks=false]{hyperref}

\iccvfinalcopy 

\def\iccvPaperID{7767} 

\ificcvfinal\fi

\begin{document}

\title{SS-IL: Separated Softmax for Incremental Learning}

\author{Hongjoon Ahn\textsuperscript{\rm 1}\thanks{Equal contribution.}, Jihwan Kwak\textsuperscript{\rm 4}\footnotemark[1], Subin Lim\textsuperscript{\rm 3}, Hyeonsu Bang\textsuperscript{\rm 1}, Hyojun Kim\textsuperscript{\rm 2} and Taesup Moon\textsuperscript{\rm 4}\thanks{Corresponding author.} \\
  \textsuperscript{\rm 1} Department of Artificial Intelligence,
  \textsuperscript{\rm 2} Department of Electronic and Electrical Engineering, \\
  \textsuperscript{\rm 3} Department of Computer Engineering, 
  Sungkyunkwan University, Suwon, Korea \\
  \textsuperscript{\rm 4} Department of Electrical and Computer Engineering, 
  Seoul National University, Seoul, Korea\\
  \texttt{\{hong0805, tnqls985, bhs1996, leopard101\}@skku.edu} \\
  \texttt{\{jihwan0508, tsmoon\}@snu.ac.kr}
}

\maketitle
\ificcvfinal\thispagestyle{empty}\fi

\begin{abstract}
   We consider class incremental learning (CIL) problem, in which a learning agent continuously learns new classes from incrementally arriving training data batches and aims to predict well on all the classes learned so far. The main challenge of the problem is the catastrophic forgetting, and for the exemplar-memory based CIL methods, it is generally known that the forgetting is commonly caused by the classification score bias that is injected due to the data imbalance between the new classes and the old classes (in the exemplar-memory). While several methods have been proposed to correct such score bias by some additional post-processing, \textit{e.g.}, score re-scaling or balanced fine-tuning, no systematic analysis on the root cause of such bias has been done. To that end, we analyze that computing the softmax probabilities by combining the output scores for all old and new classes could be the main cause of the bias. Then, we propose a new method, dubbed as Separated Softmax for Incremental Learning (SS-IL), that consists of separated softmax (SS) output layer combined with task-wise knowledge distillation (TKD) to resolve such bias. Throughout our extensive experimental results on several large-scale CIL benchmark datasets, we show our SS-IL achieves strong state-of-the-art accuracy through attaining much more balanced prediction scores across old and new classes, without any additional post-processing.
\end{abstract}

\section{Introduction}\label{sec:intro}

Incremental or continual learning, in which the agent continues to learn incremental arrival of new training data, is one of the grand challenges in artificial intelligence and machine learning. 
Such setting, which does not assume the full availability of old training data, is recently gaining more attention particularly from the perspective of real-world applications. The reason is storing all the training data, which can easily become large-scale, in one batch often becomes unrealistic for memory- and computation-constrained applications, such as mobile phones or robots. Therefore, the continuous yet effective update of the learning agent without accessing the full data received so far is indispensable. 

A viable candidate for such agent is the end-to-end learning based deep neural network (DNN) models. Following the recent success in many different applications \cite{HinLecBen15,(AutonomousDriving)Al-Qizwini,(Medical)Caruana}, the DNN-based incremental learning methods have been also actively pursued in recent years. Despite some promising results, they also possess a critical limitation: the \emph{catastrophic forgetting}, which refers to the problem that the generalization performance on the old data severely degrades after a naive fine-tuning of the model with the new data.  




In this paper, we focus on the DNN-based \emph{class} incremental learning (CIL), which we refer to learning a classifier to classify \emph{new} object classes from every incremental training data and testing the classifier on all the classes learned so far. 
Among several different proposed approaches, the exemplar-memory based ones \cite{(icarl)rebuffi17,(End2EndIL)Castro2018,(LargeScaleIL)Wu2019,(WA)Zhao2020,(IL2M)Belouadah,(scail)belouadah2020}, which allow to store small amount of training data from old classes in a separate memory, have been shown to be effective in mitigating the catastrophic forgetting. 



The main challenge of using the exemplar-memory in CIL is to resolve the severe data imbalance between the training data points for the new classes and for the old classes (in the exemplar-memory). That is, the naive fine-tuning with such imbalanced data would heavily skew the prediction scores toward the newly learned classes, hence, the accuracy for the old classes would dramatically drop, resulting in a significant forgetting. 
Recently, several state-of-the-art methods \cite{(End2EndIL)Castro2018,(LargeScaleIL)Wu2019,(WA)Zhao2020,(IL2M)Belouadah,(scail)belouadah2020} proposed to correct such score bias by some additional post-processing steps, \textit{e.g.}, score re-scaling or balanced fine-tuning, after learning the classification models.

While above mentioned methods were effective to some extent in terms of improving the accuracy, we argue that they lack systematic analyses on the main reason of such bias and that some component of their schemes, \textit{e.g.}, knowledge distillation (KD) \cite{(Distill)HintonOriolDean15}, was naively used without any proper justifications \cite{(LargeScaleIL)Wu2019,(UnlabeledIL)Lee2019,(WA)Zhao2020,(ConfCalIncre)Kang2020}. To that regard, in this paper, we first analyze the root cause of such classification score bias, then propose a method that mitigates the cause in a sensible way. Namely, we argue that the bias is injected by the fact that the softmax probability used in the ordinary cross-entropy loss is always computed by combining the output scores of \textit{all} classes, which forces the heavy penalization of the output scores for the old classes due to the data imbalance. Furthermore, we show that a naive use of the General KD (GKD) method, which also combines the output scores of \textit{all} old classes to compute the soft target, may \textit{preserve} the bias and even hurt the accuracy, if the prediction bias is already present in the model. 

To resolve above issues, we propose Separated-Softmax for Incremental Learning (SS-IL), which consists of two main components. Firstly, we devise 
\textit{separated softmax (SS)} output layer that mutually blocks the flow of the score gradients between the old and new classes, thus, mitigates the imbalanced penalization of the output probabilities for the old classes. 
Secondly, we show the \textit{Task-wise KD (TKD)} \cite{(LwF)LiHoiem16}, which also computes the soft target for the distillation in a task-separated manner, is particularly well-suited for our SS layer, since it attempts to preserve the task-wise knowledge without preserving the prediction bias that may remain across the tasks. In order to show the effectiveness of our approach, we carried out \textit{extensive} experimental validations on several large-scale CIL benchmarks with various scenarios and fairly compared our SS-IL with recent strong baselines by reproducing all of them. As a result, we convincingly show our SS-IL achieves strong state-of-the-art accuracy via adequately balancing the prediction scores across old and new classes, without \textit{any} additional post-processing.

In summary, our contribution is threefold:
\begin{compactitem}
\item We propose a novel \textit{separated softmax (SS)} layer, which prevents the old class scores from being overly penalized throughout the gradient steps.
\item We show that using GKD in CIL may preserve the bias of the model, while TKD can bring synergy when particularly combined with SS that has the same intuition.


\item We carry out extensive experimental validation of our SS-IL on several large-scale benchmarks with various CIL scenarios and fairly compared with recent, all-reproduced state-of-the-art baselines. 

\end{compactitem}

\section{Related Work}

Recently, there has been a plethora of work done to tackle the catastrophic forgetting problem in continual/incremental learning. For general continual learning, there has been three main approaches; 1) regularization-based \cite{(EWC)KirkPascRabi17,(SI)ZenkePooleGang17,(Rwalk)chaudhry2018riemannian,(UCL)ahn2019uncertainty, (AGS-CL)Jung2020}, 2) dynamic architecture-based \cite{(PNN)RusuRabiDesjSoyeKirk2016,(DEN)YoonYangLeeHwang18,(L2G)Li2019,(CPG)Hung2019}, and 3) exemplar/replay-memory based \cite{(GEM)LopezRanzato17,(A-GEM)Chaudhry2019,(ER)Chaudhry2019,(DGR)ShinLeeKimKim17,(Fearnet)kemker18,(CGANIL)Xiang2019, (DGR)ShinLeeKimKim17,(Fearnet)kemker18,(CGANIL)Xiang2019} methods. For a more thorough survey, we refer the readers to \cite{(CLSurvey)Paarisi2019}.

For CIL, in particular, the exemplar-memory based method combined with knowledge distillation (KD) has been shown to be effective. We summarize the representative recent work in the following.

\noindent\textbf{Using exemplar-memory and bias correction}
In CIL, early exemplar-memory based approaches, \eg, iCaRL \cite{(icarl)rebuffi17} and EEIL \cite{(End2EndIL)Castro2018}, have shown superior results. iCaRL classifies the examples using Nearest Mean of Exemplars (NME), and EEIL additionally exploits balanced fine-tuning, which further fine-tunes the network with a balanced training batches. Later, Javed et al. \cite{(Revisiting)Javed2018} points out that methods using exemplar-memory cause imbalanced dataset and consequently have been shown to suffer from the bias problem in the final FC layer. 
To tackle this imbalanced learning problem, several bias removal techniques have been proposed. Another balanced fine-tuning approach UW \cite{(UnlabeledIL)Lee2019} utilizes gradient scaling by weighting the losses based on the statistics of the training data. BiC \cite{(LargeScaleIL)Wu2019} corrects the bias of scores by additionally training a bias correction layer, and WA \cite{(WA)Zhao2020} corrects the biased weights in the FC layer based on the norm of each weight. Moreover, IL2M \cite{(IL2M)Belouadah} rectifies the output softmax probability and ScaIL \cite{(scail)belouadah2020} scales the classifier weights, by utilizing statistical properties of the model outputs.


\noindent\textbf{Knowledge distillation (KD)}
KD has been widely used in CIL as a popular technique to preserve and leverage the information learned from the old classes, so that the forgetting could be mitigated. However, as mentioned in the Introduction, several versions of KD have been confusingly used in different methods without proper justifications. Namely, for example, LwF \cite{(LwF)LiHoiem16}, iCaRL \cite{(icarl)rebuffi17}, and EEIL \cite{(End2EndIL)Castro2018} utilized the form of TKD, whereas  BiC \cite{(LargeScaleIL)Wu2019}, UW \cite{(UnlabeledIL)Lee2019}, and WA \cite{(WA)Zhao2020} used the form of GKD. Each method simply made its choice, and no analysis or justification on the choice is given other than some intuitive arguments (\eg, \cite{(UnlabeledIL)Lee2019} justifies GKD, but, without proper comparison or evidence.) 

Apart from above methods, LUCIR \cite{(UnifiedIL)Hou2019} and PODNet \cite{(PODNet)Arthur} considered a slightly different setting, in which an initial model is trained with large number of base classes to get useful feature representations, and they utilize the \textit{feature} distillation to preserve those representations while learning the future classes. However, we believe their setting is more limited, and we compare their performances in the \textit{pure} CIL setting to make a fair comparison with other baselines.

\section{Preliminaries}\label{sec:notation}
\subsection{Notations and problem setting}









In CIL, we assume every incrementally-arrived training data, which is often called as the incremental \textit{task}, consists of data for \textit{new} $m$ classes that have not been learned before. 
More formally, the training data for the incremental task $t$ is denoted by $\mathcal{D}_t=\{(\bm x_{t}^{(i)},  y_{t}^{(i)})\}_{i=1}^{n_{t}}$, in which $\bm x_{t}^{(i)}$, 
 $y_{t}^{(i)}$, and $n_{t}$ denote input data for task $t$, the corresponding (integer-valued) target label, and the number of training samples for the corresponding task, respectively. The total number of classes up to task $t$ is denoted by $C_t=m\cdot t$, which leads to the labeling $y_{t}^{(i)}\in\{C_{t-1}+1,\ldots,C_{t}\}\triangleq \mathcal{C}_t$.
During learning each incremental task, we assume a separate exemplar-memory $\mathcal{M}$  is allocated to store exemplar data for old classes. Namely, when learning the incremental task $t$, we store $\lfloor\frac{|\mathcal{M}|}{C_{t-1}}\rfloor$ data points from each class that are learnt until the incremental task $t-1$. Thus, as the incremental task grows, the number of exemplar data points stored for each class decreases linearly with $t$ and we assume $|\mathcal{M}|\ll n_{t}$. The total number of incremental tasks is denoted by $T$.


Our classification model consists of a feature extractor, which has the deep convolutional neural network (CNN) architecture, and the classification layer, which is the final fully-connected (FC) layer with softmax output. We denote $\bm \theta$ as the parameters for our classification model. At incremental task $t$, the parameters of the model, $\bm \theta_t$, are learned using data points in $\mathcal{D}_t\cup \mathcal{M}$. After learning, the class prediction for a given sample $\bm x_{\text{test}}$ is obtained by
\begin{eqnarray}
 \hat{y}_{\text{test}}=\arg\max_{y\in\mathcal{C}_{1:t}}z_{ty}(\bm x_{\text{test}}, \bm \theta_t),\label{eq:test}
\end{eqnarray}
in which $z_{ty}(\bm x_{\text{test}},\bm\theta_t)$ is the output score (before softmax) of the model $\bm \theta_t$ for class $y\in \{1,\ldots,C_t\}\triangleq\mathcal{C}_{1:t}$.
Namely, at test time, the final FC layers are consolidated and the prediction among all classes in $\mathcal{C}_{1:t}$ is made as if by an ordinary multi-class classifier.

\subsection{Knowledge distillation}

\begin{figure}[t]
    \centering
    \vspace{-.0in}
    \includegraphics[width=1.0\columnwidth]{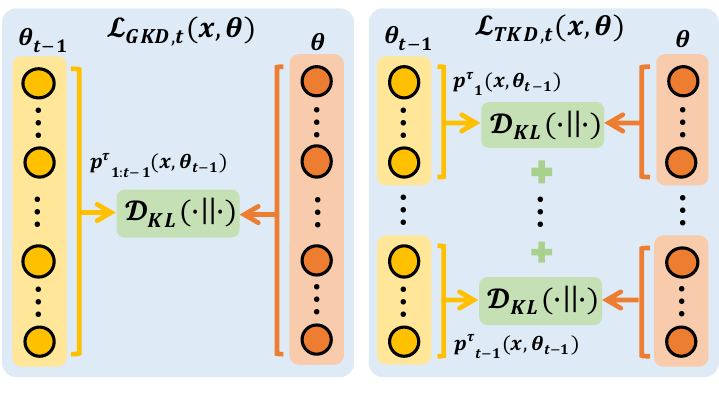}
    \vspace{-.2in}
    \caption{Illustration of $\mathcal{{L}_{\text{GKD, t}}}(\bm x, \bm \theta)$ (left) and $\mathcal{{L}_{\text{TKD, t}}}(\bm x, \bm \theta)$ (right)}
    \label{fig:TKD_GKD_illustration}
    \vspace{-.2in}
\end{figure}

As previously mentioned, the two main variations of KD used in CIL are General KD (GKD) and Task-wise KD (TKD), 
and the loss function defined for each method for learning task $t$ is as follows: for an input data $\bm x\in\mathcal{D}_t\cup\mathcal{M}$, 
\begin{align}
    &\mathcal{L}_{\text{GKD}, t}(\bm x, \bm \theta) \triangleq \mathcal{D}_{KL}(\bm p^\tau_{1:t-1}(\bm x, \bm \theta_{t-1}) \|\bm p^\tau_{1:t-1}(\bm x, \bm \theta)) \label{eq:GKD_loss} \\
    &\mathcal{L}_{\text{TKD}, t}(\bm x, \bm \theta) \triangleq \sum_{s=1}^{t-1}\mathcal{D}_{KL}(\bm p_s^\tau(\bm x, \bm \theta_{t-1}) \|\bm p_s^\tau(\bm x, \bm \theta)) \label{eq:TKD_loss},
    \vspace{-.2in}
\end{align}
in which $\mathcal{D}_{KL}(\cdot\|\cdot)$ is the Kullback-Leibler divergence, $\tau$ is a temperature scaling parameter, $\bm\theta$ are the model parameters that are being learned for task $t$, and $\bm\theta_{t-1}$ are the model parameters \textit{learned} up to task $t-1$. Furthermore, in (\ref{eq:GKD_loss}) and (\ref{eq:TKD_loss}), we define the $c$-th component of the probability vectors $\bm p_{s}^\tau(\bm x, \bm\theta)\in\Delta^m$ and $\bm p_{1:s}^\tau(\bm x,  \bm\theta)\in\Delta^{C_s}$ as 
\begin{align}
    p_{s,c}^{\tau}(\bm x, \bm \theta )=&\frac{e^{z_{sc}(\bm x,\bm \theta)/\tau}}{\sum_{k\in\mathcal{C}_s}e^{z_{sk}(\bm x,\bm \theta)/\tau}}\ \ \text{and}\nonumber\\
    p_{1:s,c}^{\tau}(\bm x, \bm \theta )=&\frac{e^{z_{sc}(\bm x,\bm \theta)/\tau}}{\sum_{k\in\mathcal{C}_{1:s}}e^{z_{sk}(\bm x,\bm \theta)/\tau}}, \nonumber
    \vspace{-.2in}
\end{align}
respectively. Namely, $\bm p_{s}^\tau(\bm x, \bm\theta)$ is the probability vector obtained by \textit{only} using the output scores for task $s$ when computing the softmax probability, and $\bm p_{1:s}^\tau(\bm x, \bm\theta)$ is the probability vector obtained by using all the output scores for tasks from $1$ to $s$ when computing the softmax probability. Thus, minimizing (\ref{eq:GKD_loss}) or (\ref{eq:TKD_loss}) will both result in regularizing the past model $\bm\theta_{t-1}$, but (\ref{eq:GKD_loss}) uses the global softmax probability across all past tasks, $\bm p^\tau_{1:t-1}(\bm x, \bm \theta_{t-1})$, while (\ref{eq:TKD_loss}) uses the task-wise softmax probabilities, $\{\bm p_s^\tau(\bm x, \bm \theta_{t-1}))\}_{s=1}^{t-1}$, obtained separately for each task. 
In recent CIL baselines, (\ref{eq:GKD_loss}) is used in \cite{(LargeScaleIL)Wu2019,(UnlabeledIL)Lee2019,(WA)Zhao2020}, and (\ref{eq:TKD_loss}) is used in \cite{(LwF)LiHoiem16,(End2EndIL)Castro2018}. The difference between (\ref{eq:GKD_loss}) and (\ref{eq:TKD_loss}) is illustrated in Figure \ref{fig:TKD_GKD_illustration}.

\section{Motivation}\label{sec:motivation}

As mentioned in the Introduction, several previous works \cite{(End2EndIL)Castro2018,(UnlabeledIL)Lee2019,(UnifiedIL)Hou2019,(LargeScaleIL)Wu2019,(IL2M)Belouadah, (WA)Zhao2020, (scail)belouadah2020} identified that the major challenge of the exemplar-memory based CIL is to resolve the classification score bias caused by the data imbalance. 
Here, we consider a simple example and give a convincing argument on why such score bias in injected as well as why a naive usage of GKD cannot fix the bias. 

Namely, first note that the ordinary cross-entropy loss for learning task $t$ used by the typical CIL methods can be expressed as 
\begin{eqnarray}
    \mathcal{L}_{\text{CE},t}((\bm x,y), \bm \theta) = \mathcal{D}_{KL}(\bm y_{1:t} \| \bm p_{1:t}(\bm x, \bm \theta)),
    \label{eq:ce_loss}
    \vspace{-.2in}
\end{eqnarray}
in which $\bm y_{1:t}$ is a one-hot vector in $\mathbb{R}^{C_t}$ that has value one at the $y$-th coordinate, and  $\bm p_{1:t}(\bm x, \bm \theta)$ is $\bm p^{\tau}_{1:t}(\bm x, \bm \theta)$ with $\tau = 1$.
Now, in order to systematically analyze the root cause of the prediction bias commonly presents in typical CIL methods, we carried out an experiment with a simple CIL method that uses the following loss 
\begin{eqnarray}
    \mathcal{L}_{\text{CE},t}((\bm x,y), \bm \theta)+\mathcal{L}_{\text{GKD}, t}(\bm x, \bm \theta)\label{eq:simple_gkd}
\end{eqnarray}
with $(\bm{x},y)\in\mathcal{D}_t\cup \mathcal{M}$ for learning task $t$. Namely, it learns the task $t$ with the cross-entropy loss while trying to preserve past knowledge by $\mathcal{L}_{\text{GKD}}$. As shown in Figure \ref{fig:Bias_motivation}, we experimented with the ImageNet dataset with $m=100$ and $|\mathcal{M}|=10k$, hence with total 10 tasks.




\subsection{Bias caused by ordinary cross-entropy}

\begin{figure}[t]
    \centering
    \vspace{-.0in}
    \includegraphics[width=1.0\columnwidth]{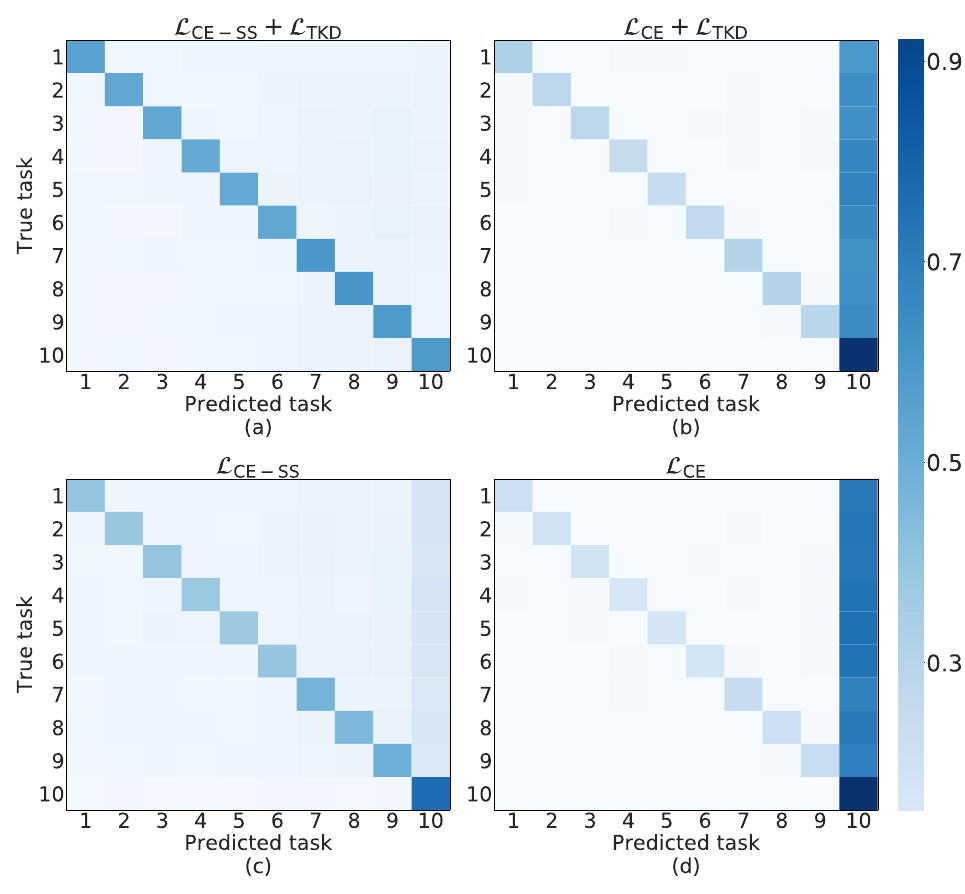}
    \vspace{-.0in}
    \caption{Left: The confusion matrix (across the tasks) based on the predictions of CIL model for test data. Right: The ratio of Top-1 predictions made by $\bm\theta_{t-1}$ on $\mathcal{D}_t$. Instead of denoting the predicted classes, we denote the tasks to which the predicted classes belong. Note that the dashed region in the right plot indicates the ratio of the latest old task, and it represents the bias on soft targets used for $\mathcal{L}_{\text{GKD}}$. All the results are on the ImageNet-1K dataset with $m=100$ and $|\mathcal{M}|=10k$.}\label{fig:Bias_motivation}
    \vspace{-.15in}
\end{figure}

\begin{figure}[t]
    \centering
    \vspace{-.0in}
    \includegraphics[width=0.8\columnwidth]{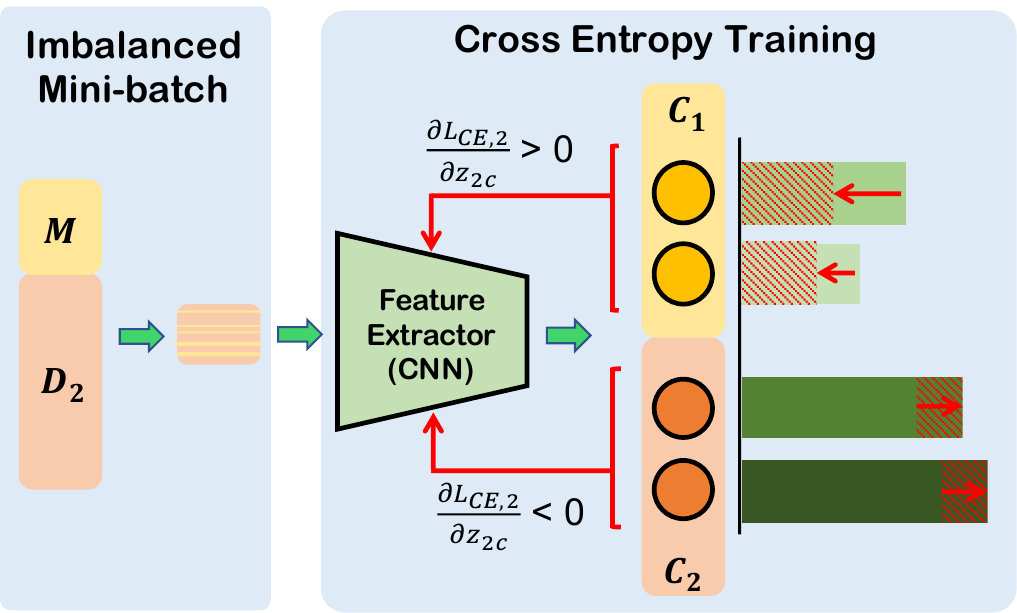}
    \vspace{-.0in}
    \caption{A toy illustration of gradient descent steps for $m=2$ and $T=2$ on imbalanced $\mathcal{D}_2\cup\mathcal{M}$. The scores for class $c\in\mathcal{C}_{1}$ continue to decrease due to the imbalanced gradient descent steps.}
    \label{fig:SS_motivation}
    \vspace{-.15in}
\end{figure}

The left plot in Figure \ref{fig:Bias_motivation} shows the confusion matrix of test samples at the task level after learning all the tasks. It clearly shows the common prediction bias; namely, most of the prediction for past tasks are overly biased toward the most recent task (task $10$). We argue that the root cause of this bias can be found in the gradient for the output score:
\begin{eqnarray}
    \frac{\partial \mathcal{L}_{\text{CE},t}((\bm x,y), \bm \theta)}{\partial z_{tc}} = p_{1:t,c}(\bm x,\bm\theta) - \mathds{1}_{\{c=y\}},\label{eq:ce_grad}
        \vspace{-.2in}
\end{eqnarray}
in which $\mathds{1}_{\{c=y\}}$ is the indicator for $c=y$. Note that since (\ref{eq:ce_grad}) is always positive for $c\neq y$, we can easily observe that when the model is being updated with data in $\mathcal{D}_t\cup \mathcal{M}$,
the classification scores for the old classes will continue to decrease during the gradient descent steps done for the abundant samples for the new classes in $\mathcal{D}_t$. Thus, 
we believe that these imbalanced gradient descent steps for the classification scores of the old classes make the significant score bias toward the new classes. The toy illustration of gradient descent steps is illustrated in Figure \ref{fig:SS_motivation}.

\subsection{Bias preserved by GKD}\label{subsec:gkd}

Now, as mentioned above, several previous works use GKD for the purpose of preserving the knowledge learned from past tasks. However, when the gradient from the cross-entropy loss causes a significant bias as mentioned in the previous section, we argue that using GKD would preserve such bias in the older model and even could hurt the performance. In $\mathcal{L}_{\text{GKD}}$ defined in (\ref{eq:GKD_loss}), $\bm p_{1:t-1}^\tau(\bm x,\bm \theta_{t-1})$ is the \textit{soft} target computed from the \textit{old} model $\bm\theta_{t-1}$, that is used for knowledge distillation. Now, Figure \ref{fig:Bias_motivation} (right) suggests that this soft target can be heavily skewed due to the bias caused by the cross-entropy learning. 
Namely, the figure shows the ratio of the \textit{tasks} among $\{1,\ldots,t-1\}$, predicted by the \textit{old} model $\bm\theta_{t-1}$ when the \textit{new} task data points $\bm x \in\mathcal{D}_{t}$ were given as input, for each new task $t$ (horizontal axis). 



%

We can observe that the predictions are overwhelmingly biased toward the most recent old task (\textit{i.e.}, task $t-1$), which is due to the bias generated during learning task $t-1$ with the cross-entropy loss. This suggests that the soft target $\bm p_{1:t-1}^\tau(\bm x,\bm \theta_{t-1})$ also would be heavily skewed toward the most recent old task (task $t-1$), hence, when it is used in GKD loss as (\ref{eq:GKD_loss}), it will preserve such bias and could highly penalize the output probabilities of the \textit{older} tasks. Hence, it could make the bias, or the \textit{forgetting} of the older tasks, more severe. 

Above two observations suggest that the main reason for the prediction bias could be to compute the softmax probability by combining the old and new tasks \textit{altogether}. Motivated by this, we propose Separated-Softmax for Incremental Learning (SS-IL) in the next section. 

\section{Main Method}\label{sec:method}



Our SS-IL mainly consists of two components, all motivated from the intuition built from the previous section: (1) Separated-Softmax (SS) output layer and (2) the Task-wise KD (TKD). We note using TKD for CIL was first proposed in LwF \cite{(LwF)LiHoiem16}, but as we show in our experiment, TKD particularly becomes powerful when combined with our SS layer. For the sake of simple explanation, at the incremental task $t$, let $\mathcal{P}_t$ denote the classes of the previous tasks ($\mathcal{C}_{1:t-1}$) and $\mathcal{N}_t$ denote the classes of the new task ($\mathcal{C}_t$).  

\vspace{.03in}

\noindent{\textit{(1) Separated-Softmax (SS) layer:}  } For $(\bm x, y)\in\mathcal{D}_t\cup \mathcal{M}$, we define a separate softmax output layer by modifying the cross-entropy loss function as 
\begin{eqnarray}
    \mathcal{L}_{\text{CE-SS},t}((\bm x, y),\bm\theta) = 
    &\mathcal{L}_{\text{CE},t-1}((\bm x,y),\bm\theta)\cdot \mathds{1}\{y\in \mathcal{P}_t\} + \nonumber\\
    &\mathcal{D}_{KL}(\bm y_{t} \| \bm p_{t}(\bm x, \bm \theta))\cdot \mathds{1}\{y\in \mathcal{N}_t\}, \label{eq:ss}
\end{eqnarray}
in which $\bm y_{t}$ stands for the one-hot vector in $\mathbb{R}^{|\mathcal{N}_t|}$ and $\bm p_{t}(\bm x, \bm \theta)$ is $\bm p_{t}^\tau(\bm x,  \bm\theta)$ with $\tau=1$. Namely, 
depending on whether $(\bm x, y)\in\mathcal{M}$  or $(\bm x, y)\in\mathcal{D}_t$, the softmax probability is computed separately by only using the output scores for $\mathcal{P}_t$ or $\mathcal{N}_t$, respectively, and the cross-entropy loss is computed separately as well. 
While (\ref{eq:ss}) is a simple modification of the ordinary cross-entropy (\ref{eq:ce_loss}), we can now observe that 
$\frac{\partial \mathcal{L}_{\text{CE-SS}}}{\partial z_{tc}}=0$ for $c\in\mathcal{P}_t$ when $(\bm x, y)\in\mathcal{D}_t$. Therefore, the gradient from the new class samples in $\mathcal{N}_t$ will not have overly penalizing effect in the classification scores for the old classes in $\mathcal{P}_t$.

\begin{figure}[t]
    \centering
    \includegraphics[width=1.0\columnwidth]{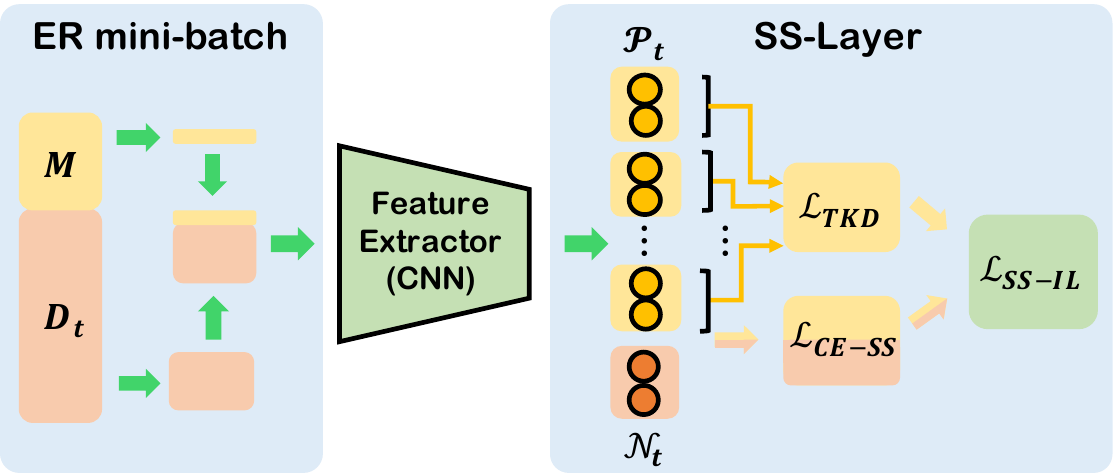}
    \caption{Illustration of SS-IL. The yellow regions represent the old classes, and red regions represent the new classes.}\label{fig:SSIL}
    \vspace{-.1in}
\end{figure}

\vspace{.03in}



\vspace{.02in}



\noindent{\textit{(2) Task-wise KD:}  }  In order to prevent preserving the bias present in GKD, we revisit the Task-wise distillation (TKD), used in LwF \cite{(LwF)LiHoiem16}. With the similar intuition as SS layer, we can easily see that it is natural to use TKD (\ref{eq:TKD_loss}), which also uses the Separated-Softmax for each task, for the knowledge distillation. That is, in TKD, since the soft targets, $\{\bm p_s^\tau(\bm x, \bm \theta))\}_{s=1}^{t-1}$, are computed only within each task, TKD will not get affected by the task-wise bias that may remain in the old model $\bm\theta_{t-1}$, as opposed to the GKD shown in Section \ref{subsec:gkd}. Hence, we can expect that TKD is particularly well-suited for the SS layer, which will be shown in our experimental results. 


\vspace{.02in}
\noindent\textit{Remark on the implementation detail: }
When random mini-batches from $\mathcal{D}_t\cup \mathcal{M}$ are naively used for the SGD updates of the model, it can also worsen the class ratio in a mini-batch towards the new classes.
Such imbalance in mini-batches is expected to downplay the updates of the model for the old classes in our SS layer, since the gradient from the first part of (\ref{eq:ss}) will be generated scarcely. Therefore, we additionally implement Experience Replay (ER) \cite{(ER)Chaudhry2019} technique that preserves the ratio of classes from old and new in a mini-batch to assure the minimum ratio of samples from $\mathcal{M}$. We empirically found that 
using ER gives more balanced prediction for SS-IL. Detailed analyses on using ER are in the Supplementary Materials.

\vspace{.03in}
\noindent \textbf{Final loss function for SS-IL: }
By combining $\mathcal{L}_{\text{CE-SS},t}$ in (\ref{eq:ss}) and $\mathcal{L}_{\text{TKD},t}$ in (\ref{eq:TKD_loss}), the overall loss for SS-IL becomes:
\begin{align}
    \mathcal{L}_{\text{SS-IL},t}((\bm x, y),\bm\theta) =  \mathcal{L}_{\text{CE-SS},t}((\bm x,  y),\bm\theta) + \mathcal{L}_{\text{TKD},t}(\bm x,\bm\theta),\nonumber \label{eq:SSIL}
    \vspace{-0.3in}
\end{align}
and the mini-batch SGD to minimize the loss is done with ER.
Figure \ref{fig:SSIL} illustrates our method, and the training algorithm is summarized in Supplementary Materials. We show in our experimental results that SS efficiently balances score between old and new classes, which, as a result, corrects the prediction bias. Finally, detailed results show that our SS-IL achieves the state-of-the-art accuracy for the various large scale benchmark datasets and in many different incremental scenarios.

\begin{table*}[t]
\small
\centering
\vspace{-.0in}
\caption{The results on various datasets and evaluation scenarios. The evaluation metrics are Average Top-1 and Top-5 accuracy. Accuracy is averaged over all the incremental tasks  (\textit{i.e.} including both initial task and incremental tasks)}\vspace{-0.0in}\label{table:Overall_results}
\resizebox{1.0\textwidth}{!}{\begin{tabular}{ccccccc}
\hline
\multicolumn{1}{c|}{$T$}     & \multicolumn{3}{c|}{$T=10$}                                                                                                                            & \multicolumn{3}{c}{$|\mathcal{M}|=10k$ (1K), $40k$ (10K)}                                                                 \\ \hline\hline
\multicolumn{1}{c|}{Dataset} & \multicolumn{1}{c|}{ImageNet-1K}                 & \multicolumn{1}{c|}{Land/mark-v2-1K}              & \multicolumn{1}{c|}{Landmark-v2-10K}             & \multicolumn{1}{c|}{ImageNet-1K}          & \multicolumn{1}{c|}{Landmark-v2-1K}       & Landmark-v2-10K      \\ \hline
\multicolumn{1}{c|}{$|\mathcal{M}|$}     & \multicolumn{1}{c|}{$5k$ / $10k$ / $20k$}        & \multicolumn{1}{c|}{$5k$ / $10k$ / $20k$}        & \multicolumn{1}{c|}{$20k$ / $40k$ / $60k$}       & \multicolumn{1}{c|}{$T = 20$ / $T = 5$}   & \multicolumn{1}{c|}{$T = 20$ / $T = 5$}   & $T = 20$ / $T = 5$   \\ \hline
                             & \multicolumn{6}{c}{Average Top-1 accuracy}                                                                                                                                                                                                                            \\ \hline\hline
\multicolumn{1}{c|}{iCaRL \cite{(icarl)rebuffi17}}   & \multicolumn{1}{c|}{47.0 / 50.5 / 53.1}          & \multicolumn{1}{c|}{37.4 / 41.1 / 44.0}          & \multicolumn{1}{c|}{23.1 / 27.2 / 29.2}          & \multicolumn{1}{c|}{44.8 / 56.2}          & \multicolumn{1}{c|}{38.1 / 45.1}          & 23.8 / 31.2          \\
\multicolumn{1}{c|}{FT \cite{(IL2M)Belouadah}}      & \multicolumn{1}{c|}{38.1 / 45.8 / 53.5}          & \multicolumn{1}{c|}{41.9 / 48.9 / 55.8}          & \multicolumn{1}{c|}{33.6 / 40.3 / 44.5}          & \multicolumn{1}{c|}{44.5 / 46.9}          & \multicolumn{1}{c|}{45.0 / 53.6}          & 38.3 / 43.1          \\
\multicolumn{1}{c|}{IL2M \cite{(IL2M)Belouadah}}    & \multicolumn{1}{c|}{41.9 / 48.4 / 55.3}          & \multicolumn{1}{c|}{42.4 / 49.2 / 55.9}          & \multicolumn{1}{c|}{34.2 / 40.7 / 44.8}          & \multicolumn{1}{c|}{45.6 / 52.4}          & \multicolumn{1}{c|}{45.2 / 54.2}          & 38.4 / 44.0          \\
\multicolumn{1}{c|}{EEIL \cite{(End2EndIL)Castro2018}}    & \multicolumn{1}{c|}{57.8 / 59.4 / 60.9}          & \multicolumn{1}{c|}{52.1 / 55.5 / 58.2}          & \multicolumn{1}{c|}{43.5 / 46.1 / 48.0}          & \multicolumn{1}{c|}{53.5 / 63.8}          & \multicolumn{1}{c|}{50.5 / 59.1}          & 41.5 / 49.8          \\
\multicolumn{1}{c|}{BiC \cite{(LargeScaleIL)Wu2019}}     & \multicolumn{1}{c|}{51.3 / 56.4 / 60.5}          & \multicolumn{1}{c|}{49.9 / 54.5 / 58.4}          & \multicolumn{1}{c|}{38.7 / 43.7 / 46.5}          & \multicolumn{1}{c|}{48.5 / 61.5}          & \multicolumn{1}{c|}{45.8 / 61.1}          & 36.3 / 50.8          \\
\multicolumn{1}{c|}{LUCIR \cite{(UnifiedIL)Hou2019}}     & \multicolumn{1}{c|}{51.0 / 53.6 / 56.5}          & \multicolumn{1}{c|}{50.5 / 53.7 / 57.3}          & \multicolumn{1}{c|}{46.2 / 49.1 / 50.9}          & \multicolumn{1}{c|}{46.8 / 61.3}          & \multicolumn{1}{c|}{48.5 / 61.0}          & \textbf{44.2} / 53.9          \\
\multicolumn{1}{c|}{PODNet \cite{(PODNet)Arthur}}     & \multicolumn{1}{c|}{52.2 / 57.5 / 60.4}          & \multicolumn{1}{c|}{-}          & \multicolumn{1}{c|}{-}          & \multicolumn{1}{c|}{48.8 / 65.5}          & \multicolumn{1}{c|}{-}          & -          \\
\multicolumn{1}{c|}{SS-IL (ours)}    & \multicolumn{1}{c|}{\textbf{63.5 / 64.5 / 65.2}} & \multicolumn{1}{c|}{\textbf{57.7 / 59.0 / 59.9} } & \multicolumn{1}{c|}{\textbf{50.1 / 51.4 / 51.9}} & \multicolumn{1}{c|}{\textbf{58.8 / 68.2}} & \multicolumn{1}{c|}{\textbf{51.4 / 64.3}} & 43.0 / \textbf{55.8} \\ \hline\hline
                             & \multicolumn{6}{c}{Average Top-5 accuracy}                                                                                                                                                                                                                            \\ \hline\hline
\multicolumn{1}{c|}{iCaRL \cite{(icarl)rebuffi17}}   & \multicolumn{1}{c|}{71.0 / 75.1 / 77.4}          & \multicolumn{1}{c|}{56.9 / 62.0 / 65.2}          & \multicolumn{1}{c|}{35.6 / 41.9 / 44.8}          & \multicolumn{1}{c|}{69.7 / 79.7}          & \multicolumn{1}{c|}{58.6 / 65.7}          & 37.8 / 46.8          \\
\multicolumn{1}{c|}{FT \cite{(IL2M)Belouadah}}      & \multicolumn{1}{c|}{66.7 / 73.3 / 78.8}          & \multicolumn{1}{c|}{62.1 / 68.5 / 74.0}          & \multicolumn{1}{c|}{49.4 / 56.7 / 60.6}          & \multicolumn{1}{c|}{71.3 / 73.0}          & \multicolumn{1}{c|}{64.6 / 72.5}          & 54.6 / 58.9          \\
\multicolumn{1}{c|}{IL2M \cite{(IL2M)Belouadah}}    & \multicolumn{1}{c|}{70.6 / 75.3 / 79.7}          & \multicolumn{1}{c|}{62.4 / 68.5 / 73.9}          & \multicolumn{1}{c|}{49.7 / 56.7 / 60.6}          & \multicolumn{1}{c|}{71.8 / 78.7}          & \multicolumn{1}{c|}{64.4 / 73.1}          & 54.3 / 59.8          \\
\multicolumn{1}{c|}{EEIL \cite{(End2EndIL)Castro2018}}    & \multicolumn{1}{c|}{81.2 / 82.0 / 83.0}          & \multicolumn{1}{c|}{72.6 / 74.9 / 76.7}          & \multicolumn{1}{c|}{60.4 / 62.6 / 64.1}          & \multicolumn{1}{c|}{77.0 / 85.3}          & \multicolumn{1}{c|}{70.3 / 77.7}          & 57.8 / 66.0          \\
\multicolumn{1}{c|}{BiC \cite{(LargeScaleIL)Wu2019}}     & \multicolumn{1}{c|}{74.4 / 78.9 / 81.8}          & \multicolumn{1}{c|}{69.2 / 73.1 / 76.1}          & \multicolumn{1}{c|}{55.5 / 60.7 / 63.3}          & \multicolumn{1}{c|}{69.4 / 84.2}          & \multicolumn{1}{c|}{63.8 / 79.3}          & 52.4 / 67.6          \\
\multicolumn{1}{c|}{LUCIR \cite{(UnifiedIL)Hou2019}}    & \multicolumn{1}{c|}{72.4 / 75.6 / 78.7}          & \multicolumn{1}{c|}{68.7 / 72.2 / 75.3}          & \multicolumn{1}{c|}{62.2 / 65.2 / 66.7}          & \multicolumn{1}{c|}{69.2 / 82.7}          & \multicolumn{1}{c|}{67.7 / 78.0}          & 60.7 / 69.3          \\
\multicolumn{1}{c|}{PODNet \cite{(PODNet)Arthur}}   & \multicolumn{1}{c|}{73.6 / 79.4 / 82.1}          & \multicolumn{1}{c|}{-}          & \multicolumn{1}{c|}{-}          & \multicolumn{1}{c|}{71.1 / 85.8}          & \multicolumn{1}{c|}{-}          & -          \\
\multicolumn{1}{c|}{SS-IL (ours)}    & \multicolumn{1}{c|}{\textbf{86.0 / 86.4 / 86.7}} & \multicolumn{1}{c|}{\textbf{78.1 / 78.8 / 79.3}} & \multicolumn{1}{c|}{\textbf{67.8 / 68.6 / 69.1}} & \multicolumn{1}{c|}{\textbf{82.9 / 88.4}} & \multicolumn{1}{c|}{\textbf{73.3 / 81.8}} & \textbf{61.8 / 72.4} \\ \hline
\end{tabular}}
\vspace{-.2in}
\end{table*}

\section{Experiments}



We believe the following two points are essential in evaluating a CIL model:
\begin{itemize}
    \addtolength\itemsep{-3mm}
    \item Evaluation on the large-scale benchmark datasets.
    \item Evaluation on diverse incremental scenarios, \eg, number of incremental tasks or size of the memory.
\end{itemize}
These points are related to the fundamentals of incremental learning considering real-world applications, which typically deal with large-scale data streams (both in data points and the number of classes) and various memory constraints. However, BiC \cite{(LargeScaleIL)Wu2019} points out that many previously proposed CIL methods fail to scale up to large-scale datasets. Also, results in \cite{(IL2M)Belouadah} and \cite{(Survey)belouadah2020} show that CIL models are sensitive to incremental conditions such as the number of incremental tasks ($T$) and exemplar-memory size ($|\mathcal{M}|$). Therefore, we
extensively compare our SS-IL with other state-of-the-art methods on two large-scale datasets (ImageNet ILSVRC 2012 \cite{(Imagenet)deng2009} and Google Landmark Dataset v2 \cite{(Landmark-v2)}) with various experimental scenarios.


For a fair comparison, we reproduced all the baselines covered in Table \ref{table:Overall_results} and compared them in 15 different CIL scenarios. Evaluation on other CIL scenario proposed in \cite{(UnifiedIL)Hou2019,(PODNet)Arthur} that pre-trains the model on large base classes and uses fixed memory size per class was also covered in Supplementary Materials. Extensive analyses of our main contribution, the Separated-Softmax (SS) layer, are carried out to show the gradient blocking effect on balancing scores between old and new classes. Lastly, in detailed analyses about the distillation methods, we show the superiority of $\mathcal{L}_{\text{TKD}}$ over $\mathcal{L}_{\text{GKD}}$. 

\subsection{Datasets and evaluation protocol}


\noindent\textbf{ImageNet and Landmark-v2}:\ \
As mentioned above, we used two large-scale benchmark datasets for our experiments, ImageNet and Google Landmark Dataset v2. ImageNet dataset consists of 1,000 classes, which has nearly 1,300 images per class. Google Landmark Dataset v2 consists of 203,094 classes, and each class has $1\sim10,247$ images. We sample 1,000 and 10,000 classes in the order of the largest number of samples per class to make two variations, and we denote each dataset as Landmark-v2-1K and Landmark-v2-10K, respectively. The number of images per class for Landmark ranges from 300 to 500 images, which inevitably provides model with imbalanced number of training data per class. By following the benchmark protocol in \cite{(icarl)rebuffi17}, we arrange the classes of each dataset in a fixed random order. We particularly stress that this is the first thorough CIL experiment on the large-scale Landmark-v2 dataset, with contains $10\times$ more classes (in Landmark-v2-10K) than previously reported results .

\noindent\textbf{Evaluation protocol}:\ \
To construct various training scenarios, we vary the total number of incremental tasks as $T=\{5,10,20\}$ which corresponds to $m=\{200,100,50\}$ in 1K datasets (ImageNet, Landmark-v2-1K) and $m=\{2000,1000,500\}$ in 10K dataset (Landmark-v2-10K), respectively.  For the exemplar-memory size, we use $|\mathcal{M}|=\{5k,10k,20k\}$ for 1K datasets and $|\mathcal{M}|=\{20k,40k,60k\}$ for 10K dataset. We use the random selection used in \cite{(Survey)belouadah2020} for constructing exemplar-memory. For the evaluation of CIL models, we use ImageNet validation set for ImageNet-1K, and we randomly select 50 and 10 images per class in Landmark-v2-1K and Landmark-v2-10K that are not in the training set, respectively. Additional explanations on dataset, evaluation protocol, and implementation detail are given in the Supplementary Materials.


\begin{figure}[t]
    \centering
    \includegraphics[width=1.0\columnwidth]{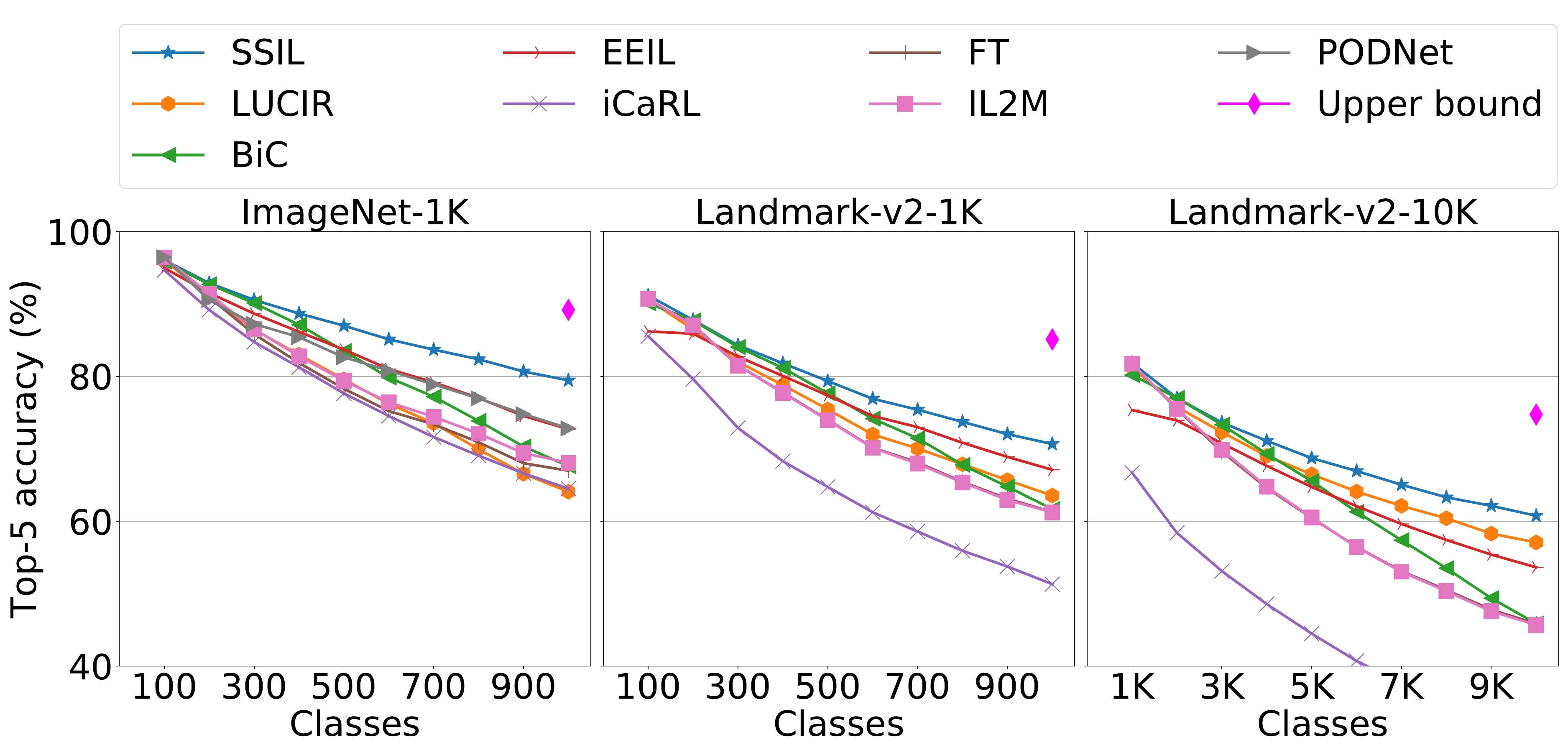}
    \vspace{-.15in}
    \caption{Top-5 accuracy results on ImageNet-1K, Landmark-v2-1K, and Landmark-v2-10K datasets for $T=10$. The exemplar size is $|\mathcal{M}|=20k$ in ImageNet-1K and Landmark-v2-1K datasets, and $|\mathcal{M}|=60k$ in Landmark-v2-10K dataset.}\label{fig:Top_5}
    \vspace{-.25in}
\end{figure}

\begin{figure*}[ht]
    \centering
    \includegraphics[width=1.0\textwidth]{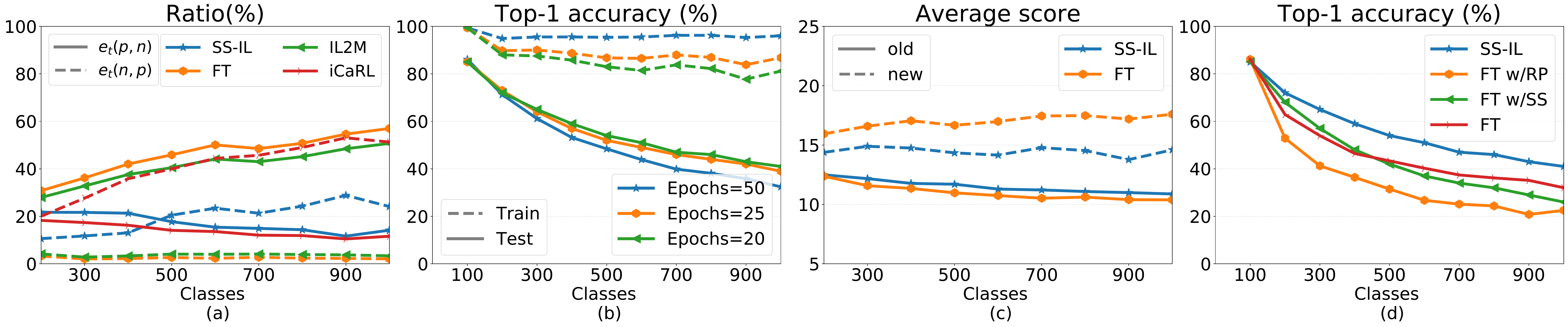}
    \vspace{-.07in}
    \caption{(a)/(b): Average classification scores for old $\&$ new classes. (c) Average softmax values for old $\&$ new classes. (d) Top-5 accuracy.}
    \label{fig:Analysis}
    \vspace{-.2in}
\end{figure*}
\subsection{Results}\label{subsec:results}
Table \ref{table:Overall_results} shows the results on Average Top-1 and Top-5 accuracy. We compare our SS-IL with iCaRL \cite{(icarl)rebuffi17}, vanilla Fine-Tuning (FT) proposed in \cite{(IL2M)Belouadah}, IL2M \cite{(IL2M)Belouadah}, EEIL \cite{(End2EndIL)Castro2018}, BiC \cite{(LargeScaleIL)Wu2019}, LUCIR \cite{(UnifiedIL)Hou2019}, and PODNet \cite{(PODNet)Arthur}. 
For each method, we used the hyperparameters reported in the original paper and run the experiments on all datasets. 
The left half of the table reports the results of fixed $T=10$ with various exemplar-memory size $|\mathcal{M}|$, and the right half shows the results of fixed $|\mathcal{M}|$ with varying $T$. We could not run PODNet on Landmark-v2 datasets due to time and memory constraints.


In Table \ref{table:Overall_results}, we observe that there is no clear winner among the baselines; EEIL tends to excel for ImageNet-1K and Landmark-v2-1K, but for Landmark-v2-10K, LUCIR achieves the highest accuracy among baselines. Furthermore, the accuracy of all the baselines drastically drops when $|\mathcal{M}|$ gets smaller. The results presented here are consistent to the conclusion of \cite{(Survey)belouadah2020} about the sensitiveness of recent CIL models to incremental scenarios and memory constraints. However, we observe that SS-IL dominates other baselines throughout almost every scenario. This indicates that SS-IL can be robustly applied to various conditions while other baselines are vulnerable to specific situation. 
Particularly, a notable characteristics of SS-IL is that the accuracy degradation of our model is minimal even when $|\mathcal{M}|$ becomes small, indicating the robustness with respect to the  exemplar-memory size.

Figure \ref{fig:Top_5} shows the overall Top-5 accuracy on each dataset with respect to the incremental task, when $|\mathcal{M}|=20k$ and $T=10$, and the tasks are denoted as classes. In this figure, we denote jointly trained approach as the Upper bound. Note that SS-IL again mostly dominates the baselines, and the performance gap over the baselines widens as the incremental task increases. Especially, in ImageNet-1K, compared with other baselines which have more performance degradation from the Upper bound, our SS-IL is less affected by catastrophic forgetting. Furthermore, we observe that iCaRL and EEIL achieve lower accuracy in the first incremental task. The Weak Nearest Exemplar Mean (NEM) classifier of iCaRL and the inefficient training schedule of EEIL could be the main reasons for such low accuracies.

\subsection{Ablation study}
In this section, we perform various detailed analyses on our SS layer to show its effect on balancing the prediction scores. 
To analyze its own strengths, we also carry out experiments that ablaties $\mathcal{L}_{\text{TKD}}$ in $\mathcal{L}_{\text{SS-IL}}$.

\begin{figure}[h]
    \centering
    \includegraphics[width=1.0\columnwidth]{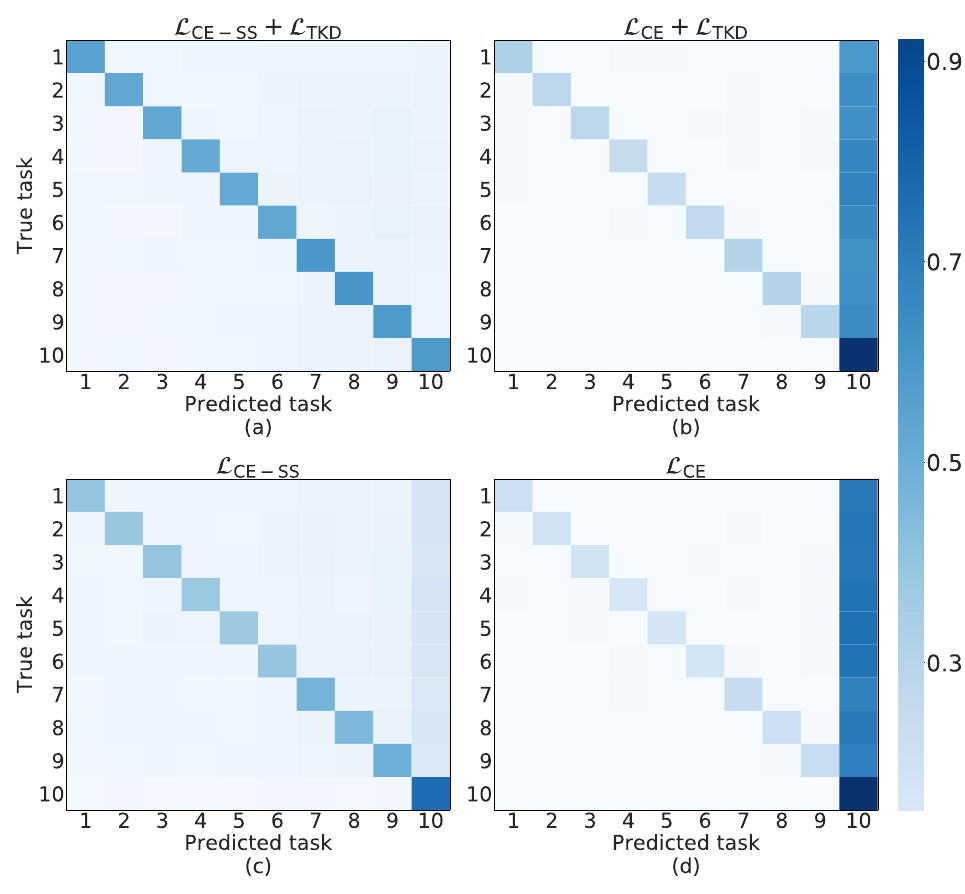}
    \vspace{-.2in}
    \caption{Confusion matrices based on the predictions of various models for test data. We denote the tasks to which the predicted class belongs.}\label{fig:confusion_matrix}
    \vspace{-.2in}
\end{figure}

Figure \ref{fig:Analysis} and \ref{fig:confusion_matrix} show the results of our analysis. In these figures, ``$\mathcal{L}_{\text{CE}}$" stands for the model that does not have both TKD and SS layer, ``$\mathcal{L}_{\text{CE}}+\mathcal{L}_{\text{TKD}}$" stands for the model that does not have SS layer, ``$\mathcal{L}_{\text{CE-SS}}$", stands for the model that only has SS, and ``$\mathcal{L}_{\text{CE-SS}}+\mathcal{L}_{\text{TKD}}$" stands for our SS-IL. We compare the four models on ImageNet-1K with $T=10$ and $|\mathcal{M}|=10k$ and analyze (a) the output scores and softmax probability values, (b) the prediction results with confusion matrix, and (c) the Top-5 accuracy.





\noindent\textbf{The output score and softmax probability}
Figure \ref{fig:Analysis} (a) and (b) show the average classification scores for the old and new classes, and Figure \ref{fig:Analysis} (c) shows the average softmax probabilities, for the old and new classes, on test samples for each incremental task $t$. When computing the average scores and softmax probabilities, we first averaged the values over the old and new classes, respectively, then we averaged them over all the test samples. 
For the softmax probability, we first normalized the scores using all the seen classes, which include the new classes. 

In Figure \ref{fig:Analysis} (a) and (b), we can observe that for ``$\mathcal{L}_{\text{CE}}+\mathcal{L}_{\text{TKD}}$" and ``$\mathcal{L}_{\text{CE}}$", the score difference between the old and new classes is significant, confirming the existence of the score bias toward the new classes. Moreover, the gap widens as the incremental task increases. For ``$\mathcal{L}_{\text{CE-SS}}+\mathcal{L}_{\text{TKD}}$" and ``$\mathcal{L}_{\text{CE-SS}}$", however, the plots for old and new classes in Figure \ref{fig:Analysis} (a) and (b) mostly overlap with each other, throughout the entire incremental learning stages. This result suggests that our SS successfully mitigates the classification score bias, without any necessary post-processing, and such balanced classification scores eventually lead to more balanced and accurate predictions.

Similarly, we also analyzed the average of the softmax probabilities normalized over all the seen classes in Figure \ref{fig:Analysis} (c). Note that since computing softmax only considers the relative difference between scores, we can compare four models without considering the magnitude of the score. From the figure, we observe that the softmax values for both old and new classes for ``$\mathcal{L}_{\text{CE-SS}}+\mathcal{L}_{\text{TKD}}$" and ``$\mathcal{L}_{\text{CE-SS}}$" are almost the same, which is consistent with the results on the output score. 


\noindent\textbf{Confusion matrix}
Figure \ref{fig:confusion_matrix} shows the confusion matrices (across the tasks) of the class predictions for the four models. In Figure \ref{fig:confusion_matrix}, we can observe that the predictions of ``$\mathcal{L}_{\text{CE-SS}}+\mathcal{L}_{\text{TKD}}$" and ``$\mathcal{L}_{\text{CE-SS}}$" are much more balanced compared to those of ``$\mathcal{L}_{\text{CE}}+\mathcal{L}_{\text{TKD}}$" and ``$\mathcal{L}_{\text{CE}}$", which have highly biased predictions. We believe above results show that TKD alone is not enough for resolving the prediction bias, and SS is essential for achieving the balanced predictions across the incremental tasks. 

\noindent\textbf{Top-5 accuracy}
Figure \ref{fig:Analysis}(d) shows the Top-5 accuracy of the four models. In this figure, as we expected, ``$\mathcal{L}_{\text{CE-SS}}+\mathcal{L}_{\text{TKD}}$" achieves the highest accuracy, and the models equipped with $\mathcal{L}_{\text{TKD}}$ outperform the models trained without it. Furthermore, we observe that ``$\mathcal{L}_{\text{CE-SS}}$" outperforms ``$\mathcal{L}_{\text{CE}}+\mathcal{L}_{\text{TKD}}$" and ``$\mathcal{L}_{\text{CE}}$", which again confirms that using SS is essential in achieving high accuracy.

\begin{figure}[t]
    \centering
    \includegraphics[width=1.0\columnwidth]{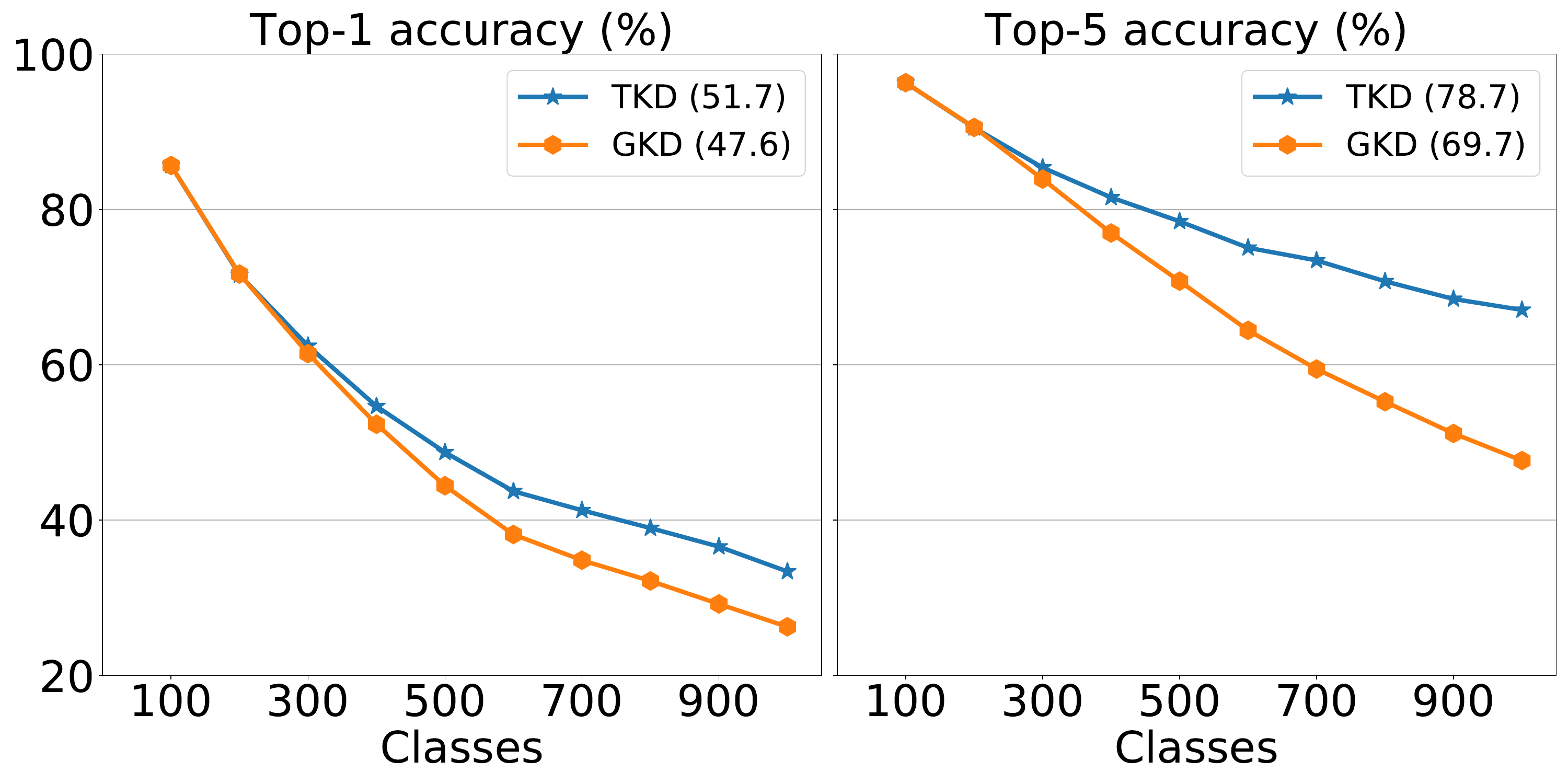}
    \vspace{-.1in}
    \caption{Top-1 accuracy (left) and Top-5 accuracy (right) for $\mathcal{L}_{\text{CE}}+\mathcal{L}_{\text{GKD}}$ and $\mathcal{L}_{\text{CE}}+\mathcal{L}_{\text{TKD}}$}\label{fig:GKD_TKD}
    \vspace{-.15in}
\end{figure}

\subsection{Analyses on KD}\label{subsec:kd}

In this section, we carry out several experiments to highlight the difference of TKD and GKD. Firstly, we evaluate the performance of models trained with TKD and GKD, respectively. Secondly, for direct comparison, we also check the accuracy of the models when each KD loss is used for the distillation from the same biased soft target. That way, we can focus only on the effect of TKD and GKD. To clarify only the effect of KD, no bias correction scheme is used for training and the same training settings are used. Training details are explained in Supplementary Materials.



\noindent\textbf{Comparison of $\mathcal{L_{\text{TKD}}}$ and  $\mathcal{L_{\text{GKD}}}$} Figure \ref{fig:GKD_TKD} shows the Top-1 and Top-5 accuracy with respect to the varying KD loss. Here, model GKD denotes the model trained with loss (\ref{eq:simple_gkd}) and model TKD stands for the one that replaces $\mathcal{L_{\text{GKD}}}$ in (\ref{eq:simple_gkd}) with  $\mathcal{L_{\text{TKD}}}$. They are trained on ImageNet-1K, $|\mathcal{M}|=10k$, and $T=10$. As shown in Figure \ref{fig:GKD_TKD}, TKD achieves much higher accuracy than GKD, while the accuracy of GKD drastically declines as the task proceeds. As shown in Figure \ref{fig:Bias_motivation}, model that does not use any bias correction method has an extreme bias and GKD may preserve such bias. On the other hand, Figure \ref{fig:GKD_TKD} shows that TKD is much less affected by the bias. \cite{(UnlabeledIL)Lee2019} argues that using TKD may miss the knowledge about discrimination between the old tasks, hence, GKD should be preferred. However, as we observe from the figure, TKD is a clear winner in terms of accuracy, since the harm of preserving the prediction bias by GKD turns out to be much greater.

\begin{figure}[t]
    \centering
    \includegraphics[width=1.0\columnwidth]{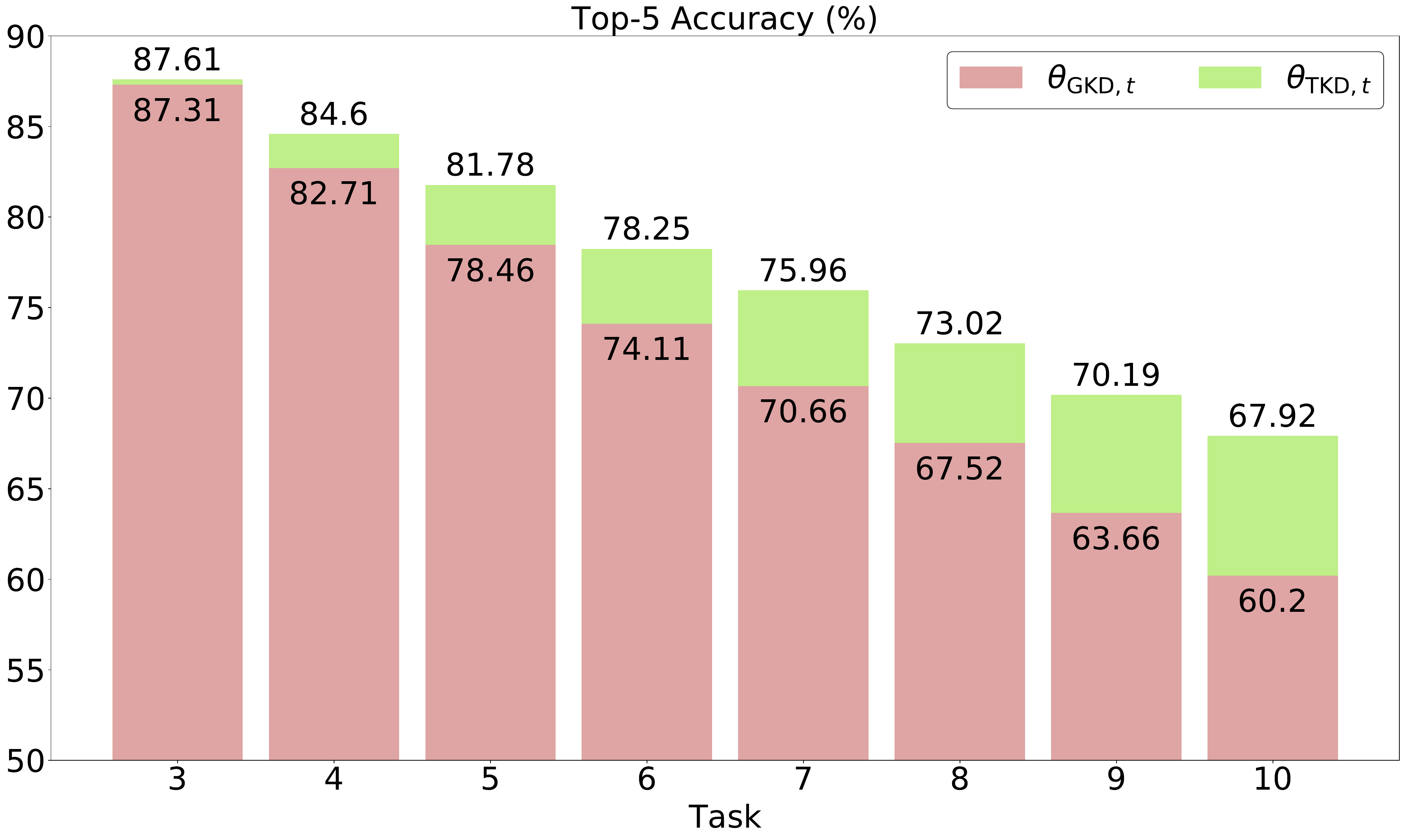}
    \vspace{-.2in}
    \caption{Top-5 accuracy for $\bm \theta_{\text{GKD},t}$ and $\bm \theta_{\text{TKD},t}$ which use $\mathcal{L_{\text{GKD}}}$ and $\mathcal{L_{\text{TKD}}}$ on same teacher model }\label{fig:global_to_local}
    \vspace{-.15in}
\end{figure}

\noindent\textbf{Comparison of $\mathcal{L}_{\text{TKD}}$ and $\mathcal{L}_{\text{GKD}}$ on the same biased teacher} In order to directly compare the behavior of $\mathcal{L_{\text{TKD}}}$ and $\mathcal{L_{\text{GKD}}}$ on the same biased soft target for distillation, we carry out another experiment with the following scenario: 
\vspace{.03in}
\begin{compactitem}
    \item[1.] Train with $\mathcal{L}_{\text{CE}}+\mathcal{L}_{\text{GKD}}$ until task $t-1$ and obtain $\bm \theta_{t-1}$.
    \item[2.] At task $t$, train $\bm \theta_{t-1}$ using two different KD losses, $\mathcal{L}_{\text{GKD}}$ and $\mathcal{L}_{\text{TKD}}$, and obtain two models, $\bm \theta_{\text{GKD},t}$ and $\bm \theta_{\text{TKD},t}$, respectively.
\end{compactitem}
\vspace{.03in}





Figure \ref{fig:global_to_local} shows the Top-5 accuracy of $\bm \theta_{\text{GKD},t}$ and $\bm \theta_{\text{TKD},t}$ at tasks $t=3,...,10$. We clearly observe that the accuracy of $\bm \theta_{\text{TKD},t}$ is always higher than that of $\bm \theta_{\text{GKD},t}$ at all tasks, which again shows that GKD preserving the score bias of $\bm \theta_{t-1}$ on $\bm x \in \mathcal{D}_t$ would be the main cause of such accuracy gap. In contrast, as in SS, TKD that utilizes the task specific separate softmax for distillation lessens the effects of the score bias. We believe this analysis is another evidence for justifying the TKD over GKD for CIL.

\section{Concluding Remarks}
We proposed SS-IL that addresses the classification score bias problem in the exemplar-memory based CIL. Our analysis suggests that ordinary softmax probability considering all the classes forces the heavy penalization of the output probabilities for the old classes, which is the main reason of the score bias. We also experimentally find that such bias is preserved by GKD and that TKD gets less affected by such bias. In our extensive experimental results, our SS-IL shows outstanding performance in most of the scenarios, and we verify that using SS can effectively balance the scores between old and new classes. 

\section*{Acknowledgments}
This work was supported in part by NRF Mid-Career
Research Program [NRF-2021R1A2C2007884] and IITP
grants [No.2019- 0-01396, Development of framework
for analyzing, detecting, mitigating of bias in AI model
and training data] [IITP-2021-2018-0-01798, ITRC Support Program], funded by the Korean government. 

\clearpage

{\small
\bibliographystyle{ieee_fullname}
\bibliography{bibfile}
}

\end{document}


\title{Supplementary Materials for \\ SS-IL: Separated Softmax for Incremental Learning}

\author{Hongjoon Ahn\textsuperscript{\rm 1}\thanks{Equal contribution.}, Jihwan Kwak\textsuperscript{\rm 4}\footnotemark[1], Subin Lim\textsuperscript{\rm 3}, Hyeonsu Bang\textsuperscript{\rm 1}, Hyojun Kim\textsuperscript{\rm 2} and Taesup Moon\textsuperscript{\rm 4}\thanks{Corresponding author.} \\
  \textsuperscript{\rm 1} Department of Artificial Intelligence,
  \textsuperscript{\rm 2} Department of Electronic and Electrical Engineering, \\
  \textsuperscript{\rm 3} Department of Computer Engineering, 
  Sungkyunkwan University, Suwon, Korea \\
  \textsuperscript{\rm 4} Department of Electrical and Computer Engineering, 
  Seoul National University, Seoul, Korea\\
  \texttt{\{hong0805, tnqls985, bhs1996, leopard101\}@skku.edu} \\
  \texttt{\{jihwan0508, tsmoon\}@snu.ac.kr}
}

\maketitle

This material specifies data configuration in CIL scenarios and model implementation details on SS-IL and other baselines. It also includes additional experimental results on ER mini-batch ablation study and other CIL scenarios. Additional CIL setting \cite{(UnifiedIL)Hou2019, douillard2020podnet} consists of large initial base classes and limited memory usage per class. Lastly, model performance with respect to the incremental task is reported to show the overall behavior.

For brevity, we use the term \textit{large base} for scenarios that gives 50\% of total classes as initial task and \textit{base} for those which consider fixed number of classes across all tasks. Also, \textit{memory per class} is used for exemplar-memory constraint allowing only constant number of samples per classes and \textit{fixed memory} for alleviated memory usage to fully store samples from seen classes. For example, Table 1 (Manuscript) corresponds to \textit{base} and \textit{fixed memory} setting. As a result, we show comprehensive results of the models in 4 different settings and 15 different conditions within each setting, which results in 60 scenarios in total.

\section{Datasets and evaluation protocol}


\noindent\textbf{ImageNet}:\ \
ILSVRC 2012 dataset consists of 1,000 classes, which has nearly 1,300 images per class. By following the benchmark protocol in \cite{(icarl)rebuffi17}, we arrange the classes of each dataset in a fixed random order. For the evaluation of CIL models, we use ILSVRC 2012 validation set for testing.

In Table 1 (Manuscript), we experiment with varied total number of incremental tasks, $T=\{5,10,20\}$, which corresponds to $m=\{200,100,50\}$ per task, and for the exemplar-memory size, we use $|\mathcal{M}|=\{5k,10k,20k\}$. When constructing exemplar-memory, we use Random selection \cite{(Survey)belouadah2020} for fixed memory setting, which simply samples random data from old classes. As exemplars from the new classes are randomly selected, it is required to delete exemplars from the old classes. In order to maintain balanced number of exemplars across all the old classes, classes that have more exemplars are selected and exemplars for the corresponding class become more likely to be removed. By doing so, difference of the number of samples across the classes is at most 1. 

In growing memory setting, we use Ring buffer approach proposed in \cite{(ER)Chaudhry2019}. We stored a constant number of samples per old class, which we denote $|\mathcal{M}_{per}|=\{5, 10, 20\}$. Thus, the number of samples stored in the memory grows as new tasks sequentially arrive. In large base setting, we first train the model with 50\% of total classes and incrementally learn additional classes per task which corresponds to $m=\{100,50,25\}$. Here, we also compare the experimental results of two exemplar-memory managing approach : growing memory and fixed memory. 





\noindent\textbf{Landmark-v2}:\ \
Google Landmark Dataset v2 consists of 203,094 classes, and each class has $1\sim10,247$ images. Since the dataset is highly imbalanced, we sample 1,000 and 10,000 classes in the order of largest number of samples per class. We denote Landmark-v2 dataset with 1,000 and 10,000 classes as Landmark-v2-1K and Landmark-v2-10K, respectively. After sampling the classes, we arrange the classes in a fixed random order. For evaluation, we randomly select 50 and 10 images per each class in Landmark-v2-1K and Landmark-v2-10K that are not in the training set for testing.

Since Landmark-v2-1K consists of same number of classes with ImageNet, all the figures regarding memory size ($|\mathcal{M}|$ or $|\mathcal{M}_{per}|$) and task numbers($T$) are similar with ImageNet. However, in Landmark-v2-10K which is composed of $10,000$ classes, the number of classes in each task is changed to  $\{2000,1000,500\}$ when we set $T=\{5,10,20\}$ in base setting and to $\{1000,500,250\}$ in large base setting. For exemplar-memory size, we use $|\mathcal{M}|=\{20k,40k,60k\}$ in fixed memory and $|\mathcal{M}_{per}| =\{2, 4, 6\}$. in growing memory setting.


\begin{table*}[t]
\small
\centering  
\vspace{-.0in}
\caption{Hyper-parameters for all methods. Details of post-processing implementations on each method are in Section \ref{subsec:baselines}}\vspace{-0.0in}\label{table:Hyperparameters}
\resizebox{1.0\textwidth}{!}{\begin{tabular}{c|c|c|c|c|c|c|ccc}
\cline{2-7}
\multicolumn{1}{l|}{}                           & \multicolumn{3}{c|}{Training}  & \multicolumn{3}{c|}{Post-processing} &                                    & \multicolumn{1}{l}{}            & \multicolumn{1}{l}{}                        \\ \hline
\multicolumn{1}{|c|}{Methods / Hyper-parameters} & Epochs  & LR   & LR schedule   & Epochs    & LR       & LR schedule   & \multicolumn{1}{c|}{LR decay rate} & \multicolumn{1}{c|}{Batch size} & \multicolumn{1}{c|}{Other hyper-parameters} \\ \hline
\multicolumn{1}{|c|}{iCaRL}                     & 60      & 0.1  & 20 30 40 50   & -         & -        & -             & \multicolumn{1}{c|}{0.2}           & \multicolumn{1}{c|}{128}        & \multicolumn{1}{c|}{-}                      \\
\multicolumn{1}{|c|}{FT}                        & 100    & 0.1  & 40 80         & -         & -        & -             & \multicolumn{1}{c|}{0.1}           & \multicolumn{1}{c|}{128}        & \multicolumn{1}{c|}{-}                      \\
\multicolumn{1}{|c|}{IL2M}                      & 100     & 0.1  & 40 80         & -         & -        & -             & \multicolumn{1}{c|}{0.1}           & \multicolumn{1}{c|}{128}        & \multicolumn{1}{c|}{-}                      \\
\multicolumn{1}{|c|}{EEIL}                      & 40      & 0.1  & 10 20 30      & 30        & 0.01     & 10 20         & \multicolumn{1}{c|}{0.1}           & \multicolumn{1}{c|}{128}        & \multicolumn{1}{c|}{-}                      \\
\multicolumn{1}{|c|}{BiC}                       & 100     & 0.1  & 30 60 90      & 200       & 0.001    & 60 120 180    & \multicolumn{1}{c|}{0.1}           & \multicolumn{1}{c|}{256}        & \multicolumn{1}{c|}{-}                      \\
\multicolumn{1}{|c|}{LUCIR}                     & 90      & 0.1  & 30 60         & -         & -        & -             & \multicolumn{1}{c|}{0.1}           & \multicolumn{1}{c|}{128}        & \multicolumn{1}{c|}{$m=0.5, K=2, \lambda_{base}=10$}              \\
\multicolumn{1}{|c|}{PODNet}                    & 90      & 0.05 & -             & 20        & 0.01     & -             & \multicolumn{1}{c|}{0.1}           & \multicolumn{1}{c|}{64}         & \multicolumn{1}{c|}{$\lambda_{c}=8,\lambda_{f}=10$}                   \\
\multicolumn{1}{|c|}{SS-IL}                     & 100     & 0.1  & 40 80         & -         & -        & -             & \multicolumn{1}{c|}{0.1}           & \multicolumn{1}{c|}{128}        & \multicolumn{1}{c|}{See details}            \\ \hline
\end{tabular}}
\end{table*}

\begin{figure*}[h]
    \centering
    \includegraphics[width=1.0\textwidth]{figures/ER_ablation.pdf}
    \caption{(a) Average classification score for old and new classes. (b) and (c) confusion matrix, (d) Top-5 accuracy.}
    \label{fig:ER_ablation}
\end{figure*}

\section{Implementation details}\label{sec:Implementation details}
\subsection{Baselines and SS-IL}\label{subsec:baselines}

All the baselines use the Resnet-18 \cite{(Resnet)HEZhang15} architecture and are implemented using PyTorch framework \cite{(PyTorch)Paszke2017}. The weight decay is set to 0.0001 and stochastic gradient descent (SGD) is used with momentum 0.9. Softmax scaling parameter($\tau$) used for distillation in Eq.(2) and Eq.(3) ( Manuscript) is set to 2. Details of the hyper-parameters are summarized in Table \ref{table:Hyperparameters} and additional explanations on each baseline model are written in this section.

We planned to consider WA \cite{(WA)Zhao2020} as one of our baselines for comparison. However, it was unable to compare our method with WA, since it did not publish its official code and reproducing its algorithm was unfeasible. Including SS-IL and all the other baselines, the code implementations will be publicly available.


\noindent\textbf{iCaRL \cite{(icarl)rebuffi17}}:\ \
Considering the implementations proposed in \cite{(Revisiting)Javed2018}, instead of binary cross entropy, multi-class cross entropy loss is used for both classification loss and KD loss. 

\noindent\textbf{FT and IL2M \cite{(IL2M)Belouadah}}:\ \
After initial task, the training epochs and the learning rate decay schedule are divided by 4, \textit{i.e.} 25 epochs and learning rate decay at 10 and 20 epoch.


\noindent\textbf{EEIL \cite{(End2EndIL)Castro2018}}:\ \
During post-processing, balanced fine-tuning is applied to both the feature extractor and the classifier. Same with iCarL, muti-class cross entropy is used for both classification loss and KD loss.

\noindent\textbf{BiC \cite{(LargeScaleIL)Wu2019}}:\ \
After training, BiC additionally trains bias correction layer using auxiliary validation set. We used 9:1 split for train:val split, which is reported to be the best choice in the original paper.

\noindent\textbf{LUCIR \cite{(UnifiedIL)Hou2019}}:\ \
LUCIR indicates the ``CNN" based method which shows better performance in large scale dataset compared with ``NEM" based method. Also, additional balanced fine-tuning for post-processing was not implemented, since its effect is insignificant according to \cite{(UnifiedIL)Hou2019}.


\noindent\textbf{PODNet \cite{(PODNet)Arthur}}:\ \
For the classifier, $10$ proxies are used, and for faster convergence, NCA loss is used by setting margin and scale to $0.6$ and $1$. For post-processing, balanced fine-tuning is performed on the output classifier.

\noindent\textbf{SS-IL (ours)}
The batch size used for $\mathcal{D}_t$, \textit{i.e.} $N_{\mathcal{D}_t}$, is 128, and we use different replay batch size, $N_{\mathcal{M}}$, depending on the number of different incremental tasks; \ie,  $N_{\mathcal{M}}=16 / 32 / 64$ for $T=20 / 10 / 5$, respectively. Thus, the ratio of $N_{\mathcal{D}_t}$ over $N_{\mathcal{M}}$ is $8 /4/2$, respectively.

\subsection{Details on KD analysis models}\label{subsec:models on kd analysis}

For a fair comparison, two models used in Section 6.4 (Manuscript) are implemented on the same experimental conditions and differ in the KD loss term ($\mathcal{L}_{\text{TKD}}$ and $\mathcal{L}_{\text{GKD}}$). We use the Resnet-18 \cite{(Resnet)HEZhang15} architecture and the stochastic gradient descent (SGD) with momentum $0.9$. The number of epochs for training incremental task is 100. The learning rate starts at 0.1 and is divided by 10 at 40 and 80 epochs. The weight decay is set to 0.0001 and the batch size is 128. Softmax scaling parameter($\tau$) used for distillation is set to 2. It was trained on base and fixed memory setting which chooses Random selection for exemplar-memory construction.

\section{Algorithm}\label{sec:algorithm}

Algorithm \ref{alg:Algorithm} shows the overall training mechanism of our SS-IL. 

\begin{algorithm}[h]
\caption{Separated Softmax for Incremental Learning (SS-IL)}
\begin{algorithmic}
\REQUIRE $\{\mathcal{D}_t\}_{t=1}^T$: Training dataset
\REQUIRE $\mathcal{M}\leftarrow \{\}$: Memory buffer
\REQUIRE $E$: The number of epochs per task.
\REQUIRE $N_{\mathcal{D}_t}, N_{\mathcal{M}}$: Training \& replay batch sizes
\REQUIRE $\alpha$: Learning rate
\REQUIRE $\bm\theta$ : Network parameters
\STATE \textcolor{blue}{\# Start class incremental learning}
\STATE Randomly initialize $\bm\theta$
\FOR{$t=1,...,T$}
\FOR{$e=1,...,E$} 
\STATE \textcolor{blue}{\# Sample a mini-batch of size $N_{\mathcal{D}_t}$}
\FOR{$B_{\mathcal{D}_t} \sim \mathcal{D}_t$} 
\STATE \textcolor{blue}{\# Sample a mini-batch of size $N_{\mathcal{M}}$}
\STATE $B_{\mathcal{M}} \sim \mathcal{M}$ 
\STATE $\mathcal{L}_t(\bm \theta) = \sum_{(\bm x, \bm y)\in B_{\mathcal{D}_t} \cup B_{\mathcal{M}}} \mathcal{L}_{\text{SS-IL},t}((\bm x, \bm y), \bm \theta)$
\STATE $\bm\theta \leftarrow \bm\theta - \frac{\alpha}{N_{\mathcal{D}_t}+N_{\mathcal{M}}} \cdot \nabla_{\bm\theta}\mathcal{L}_t(\bm \theta)$ 
\ENDFOR
\ENDFOR
\STATE $\mathcal{M}\leftarrow$ UpdateMemory($\mathcal{D}_t$, $\mathcal{M}$)
\ENDFOR
\end{algorithmic}
\label{alg:Algorithm}
\end{algorithm}


%


\section{Analysis on ER mini-batch}

\begin{table}[h]
\small
\centering
\vspace{-.0in}
\caption{Results on ImageNet-1K with varying $N_{\mathcal{M}}$ and $T$.}\vspace{-0.0in}\label{table:ER_mini_batch}
\resizebox{1.0\columnwidth}{!}{\begin{tabular}{cc|c}
\hline
\multicolumn{1}{c|}{$T$/$N_{\mathcal{M}}$} & 16 / 32 / 64           & 16 / 32 / 64           \\ \hline
                         & Average Top-1 accuracy & Average Top-5 accuracy \\ \hline
\multicolumn{1}{c|}{20}  & 58.8 / 59.0 / 58.9     & 82.9 / 82.6 / 82.4     \\
\multicolumn{1}{c|}{10}  & 54.3 / 64.5 / 68.2     & 86.6 / 86.4 / 86.0     \\
\multicolumn{1}{c|}{5}   & 68.4 / 68.4 / 68.2     & 88.8 / 88.6 / 88.4     \\ \hline
\end{tabular}}
\end{table}

In this section, we carry out analyses on ER mini-batch for ImageNet-1K with $T=10$ and $|\mathcal{M}|=10k$. Figure \ref{fig:ER_ablation} shows the ablation study results on ER mini-batch. Note that ``SS-IL w/o ER" stands for SS-IL without ER mini-batch. In Figure \ref{fig:ER_ablation} (a), similarly as the results shown in Manuscript, due to the effectiveness of SS, ``SS-IL w/o ER" also has balanced output scores. By comparing Figure \ref{fig:ER_ablation} (b) and (c), ``SS-IL" shows little more balanced predictions, and as a result, ``SS-IL" shows minute increase in the final task. Though the effect of using ER mini-batch is marginal, to get more balanced prediction and well performing results, we use it as an additional technique.

Table \ref{table:ER_mini_batch} shows the results on Average Top-1 and Top-5 accuracy with respect to varying ER mini-batch size, $N_{\mathcal{M}}$, and the total number of incremental tasks, $T$. From the table, we observe that no matter what $N_{\mathcal{M}}$ is being used, the accuracy differences are negligible. This indicates that using ER mini-batch is effective regardless of the ratio between old and new class samples in the mini-batch, if the old class examples are guaranteed to some extent.

\section{Additional results}

\subsection{Additional CIL scenarios}\label{subsec:growing memory}

Table \ref{table:growing_table}, \ref{table:largebase_table}, and \ref{table:both_table} report Average Top-1 and Top-5 accuracy in ImageNet-1K and Landmark-v2-1K. Each table represents different CIL setting depending on additional memory constraint and base class quantity. Namely, CIL scenarios in Table \ref{table:largebase_table} and Table \ref{table:both_table} assume recently proposed large base setting. Also, Table \ref{table:growing_table} and Table \ref{table:both_table} assume growing memory setting which has more strict memory constraint. Baseline models that show comparable results among those in Table 1(Manuscript) are selected for evaluation. Due to time and memory limitations, we are currently unable to report all the results for Landmark-v2-10K dataset, particularly for PODNet \cite{(PODNet)Arthur}. 
We will make sure to update the remaining results as soon as possible in an arXiv version.

In Table \ref{table:growing_table}, we clearly observe that our SS-IL is superior to other methods for the hard memory constraint setting. Results in $T=10$ and $|\mathcal{M}_{per}|=\{5,10,20\}$ show that SS-IL with $|\mathcal{M}_{per}|=5$ even excel other models that use two or four times more images per class. Also, compared to results in Table 1 (Manuscript), SS-IL shows much small Top-1 accuracy drop caused by most of the cases, \textit{e.g}, SS-IL(1.3\%$\downarrow$), EEIL(8.9\%$\downarrow$), BIC(5.5\%$\downarrow$), LUCIR(3.8\%$\downarrow$), PODNet(3.7\%$\downarrow$) at ImageNet-1K $T=10$ and $|\mathcal{M}|=20\text{K}$, which corresponds to ImageNet-1K $T=10$ and $|\mathcal{M}_{per}|=20$. These results show SS-IL's strong robustness in memory conditions which in turn leads to highest performance in most CIL scenarios.





 In Table \ref{table:largebase_table} and \ref{table:both_table}, we clearly observe that SS-IL outperforms strong baselines in large base setting (LUCIR, PODNet) as well on many scenarios. Note that unlike LUCIR and PODNet that utilize large base setting tailored algorithm, SS-IL does not assume any CIL scenario in learning objective and achieves as much or even better performance. SS-IL also shows no significant difference in accuracy when growing memory setting is adapted in Table \ref{table:both_table}. 
 
\subsection{Overall Top-1 and Top-5 accuracy}\label{subsec:overall top-1 and top-5}

\begin{figure*}[t]
    \centering
    \subfigure[$T=10$ and $|\mathcal{M}|=5k$(1K), $20k$(10K)]{
    \includegraphics[width=0.6\textwidth]{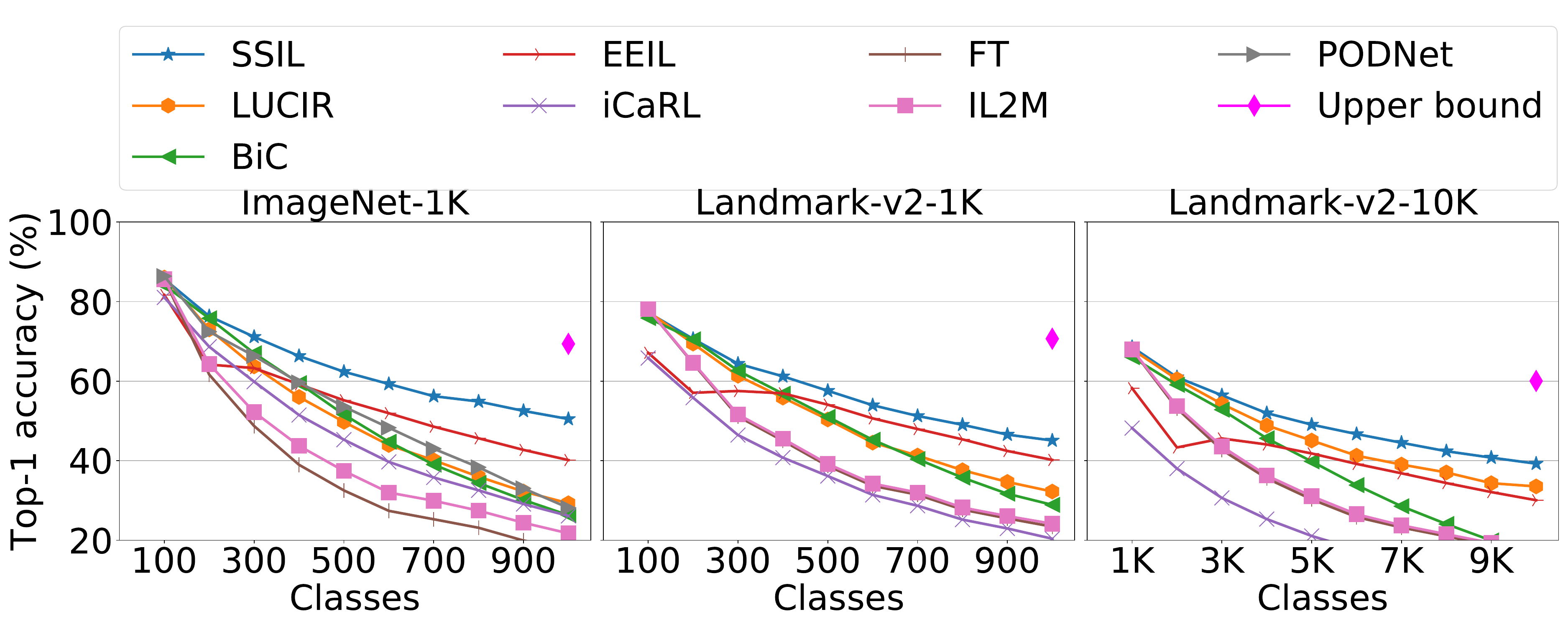}}
    \subfigure[$T=10$ and $|\mathcal{M}|=10k$(1K), $40k$(10K)]{
    \includegraphics[width=0.6\textwidth]{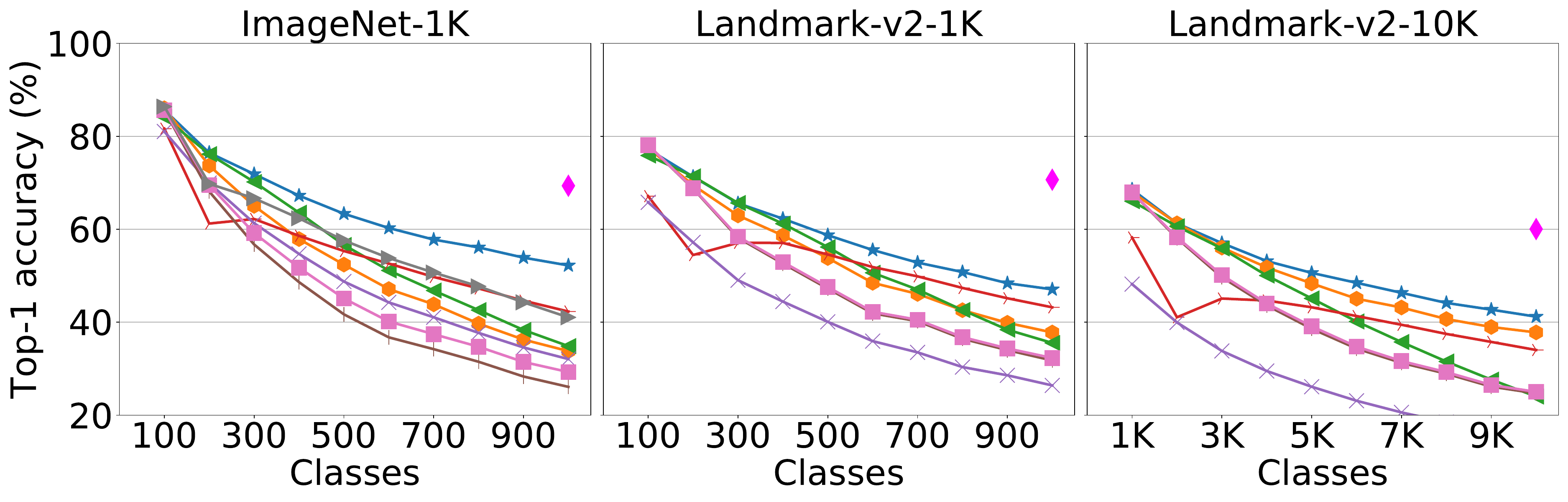}}
    \subfigure[$T=10$ and $|\mathcal{M}|=20k$(1K), $60k$(10K)]{
    \includegraphics[width=0.6\textwidth]{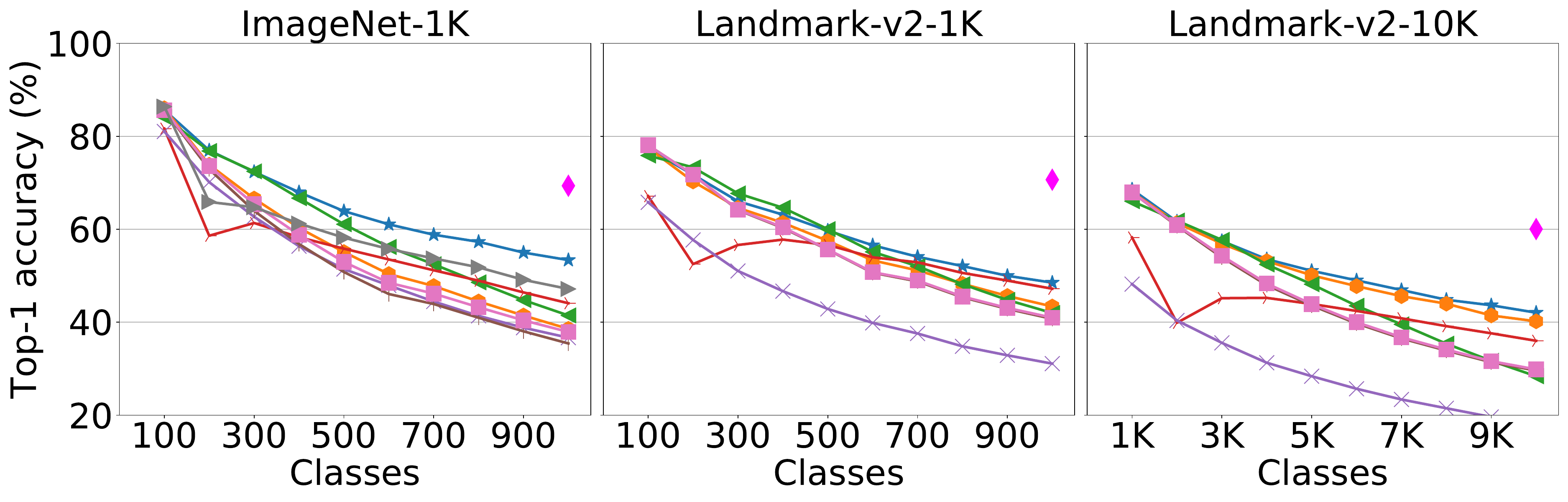}}
    \subfigure[$T=20$ and $|\mathcal{M}|=10k$(1K), $40k$(10K)]{
    \includegraphics[width=0.6\textwidth]{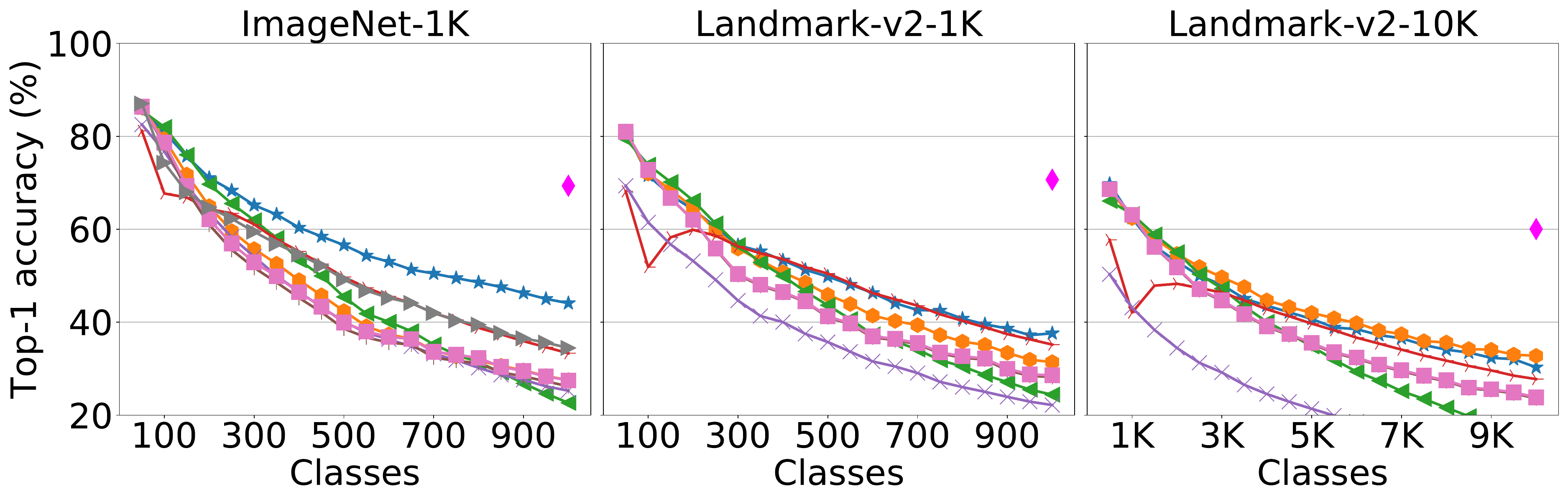}}
    \subfigure[$T=5$ and $|\mathcal{M}|=10k$(1K), $40k$(10K)]{
    \includegraphics[width=0.6\textwidth]{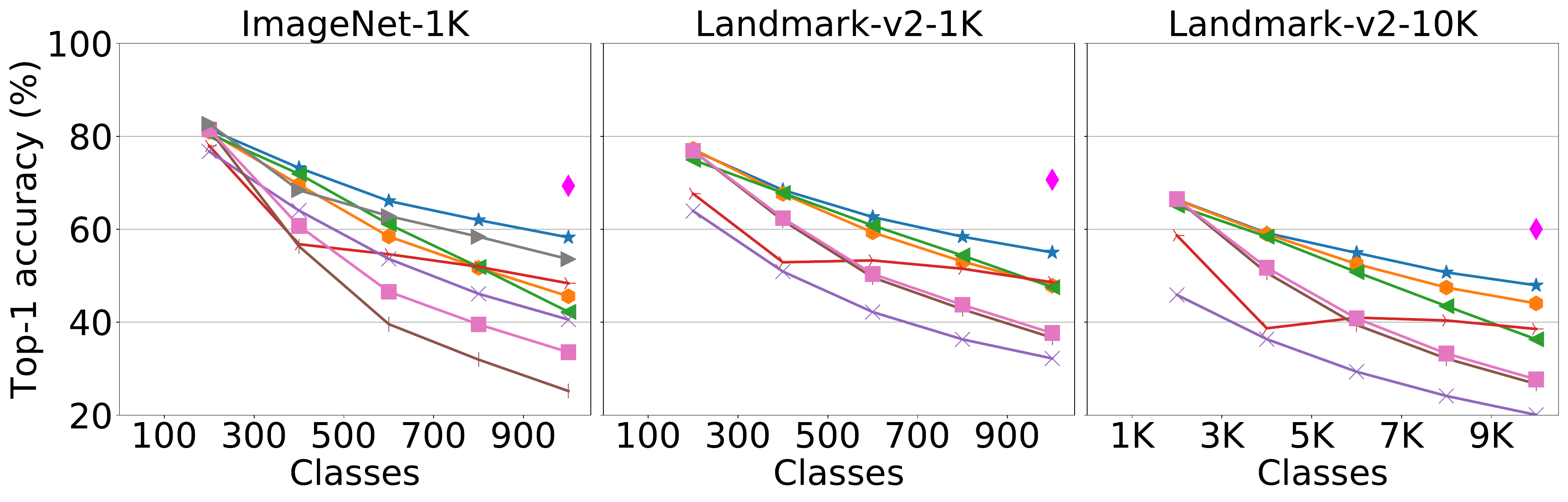}}
    \caption{Top-1 accuracy on ImageNet-1K, Landmark-1K and, Landmark-10K.}\label{fig:results_top_1}
\end{figure*}

\begin{figure*}[t]
    \centering
    \subfigure[$T=10$ and $|\mathcal{M}|=5k$(1K), $20k$(10K)]{
    \includegraphics[width=0.6\textwidth]{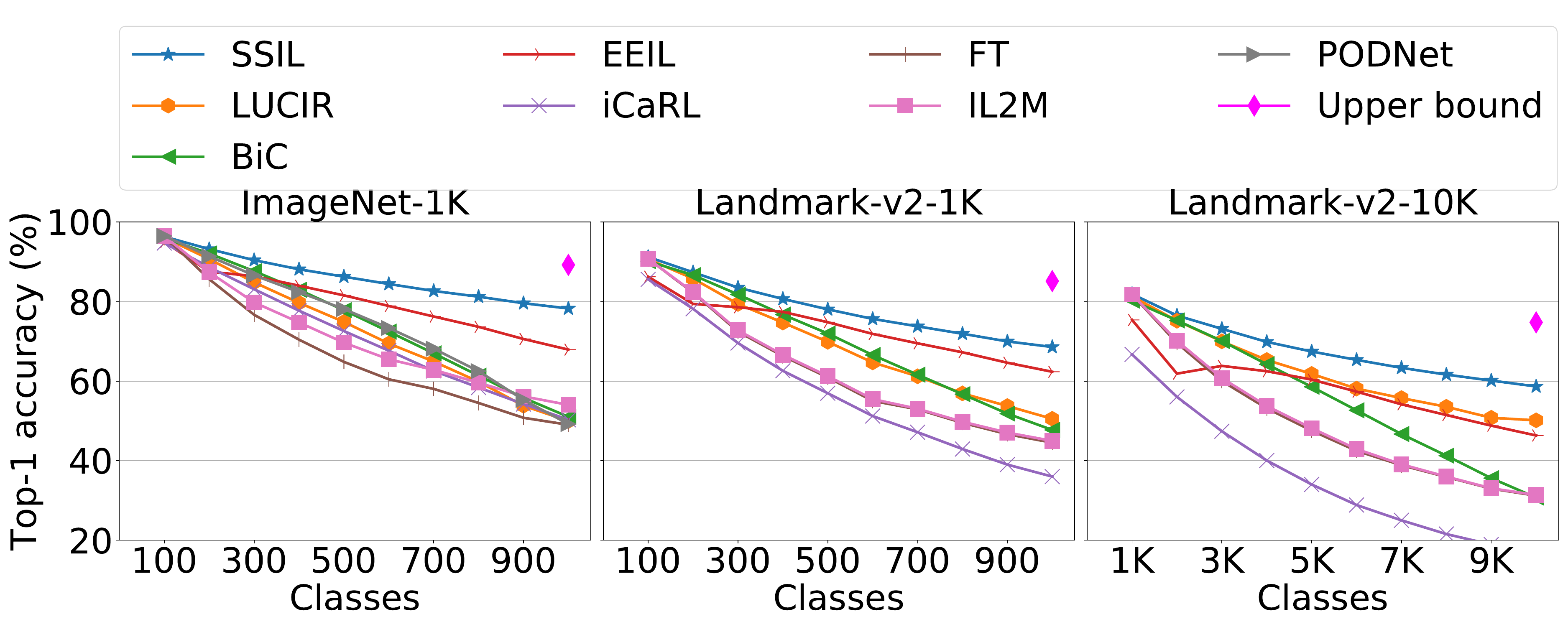}}
    \subfigure[$T=10$ and $|\mathcal{M}|=10k$(1K), $40k$(10K)]{
    \includegraphics[width=0.6\textwidth]{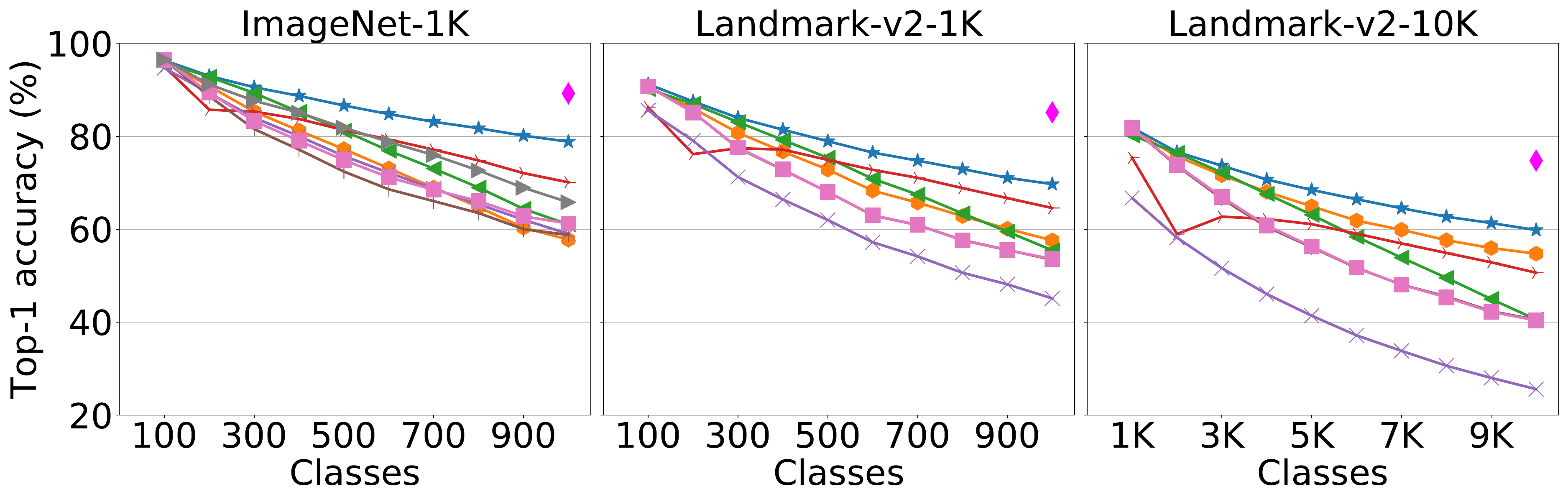}}
    \subfigure[$T=10$ and $|\mathcal{M}|=20k$(1K), $60k$(10K)]{
    \includegraphics[width=0.6\textwidth]{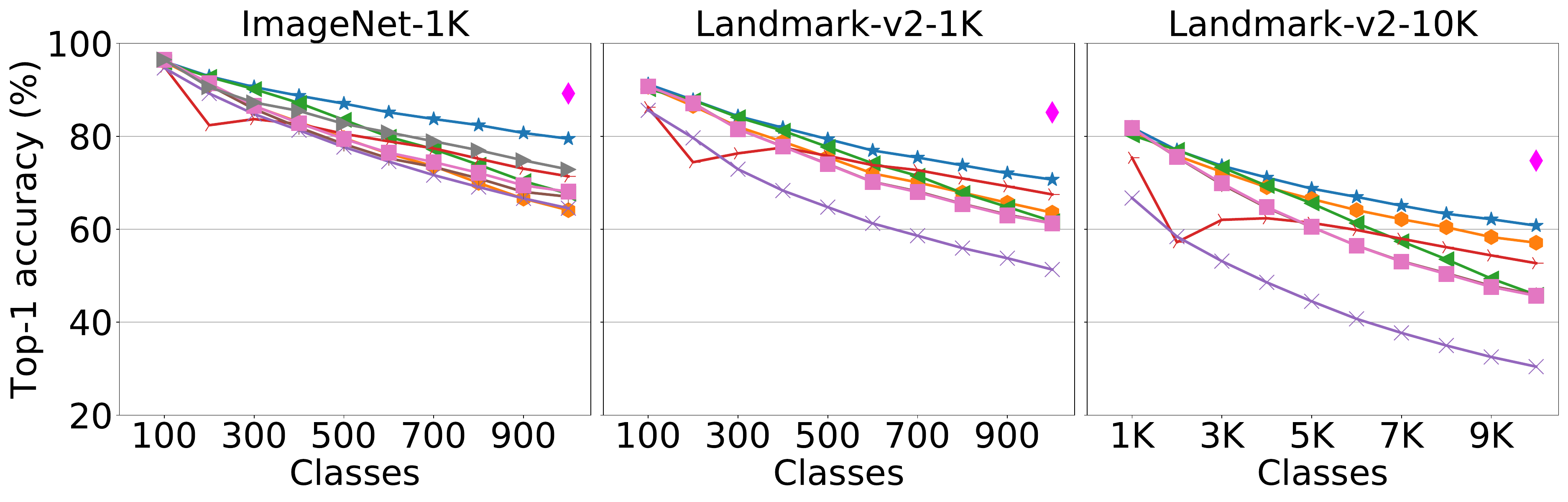}}
    \subfigure[$T=20$ and $|\mathcal{M}|=10k$(1K), $40k$(10K)]{
    \includegraphics[width=0.6\textwidth]{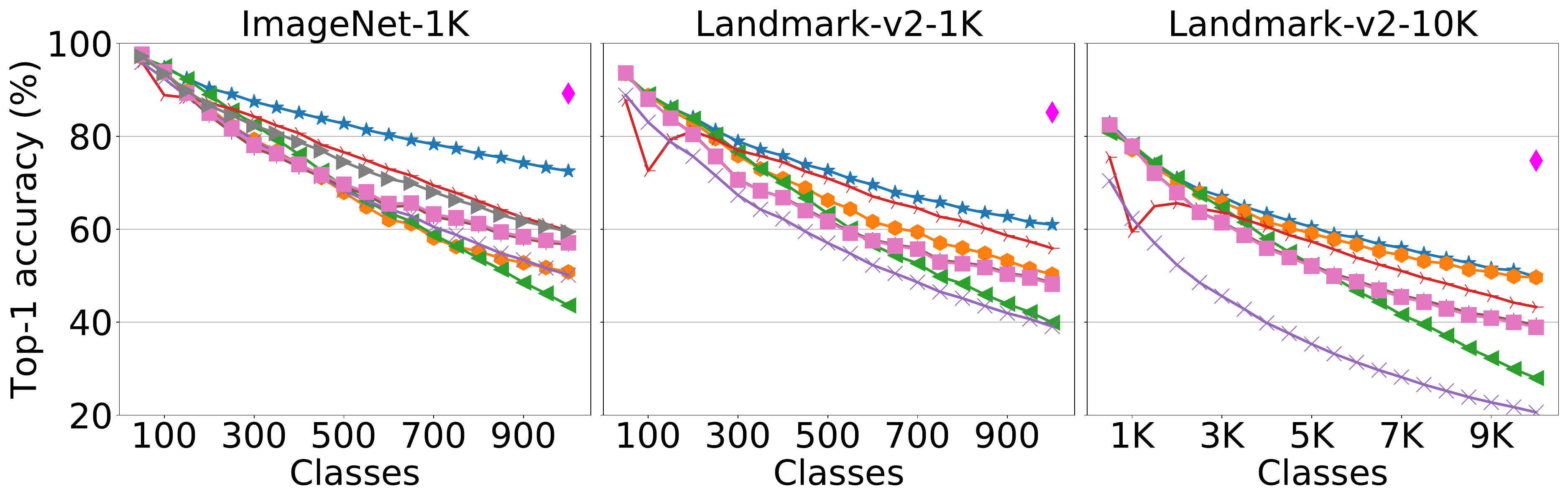}}
    \subfigure[$T=5$ and $|\mathcal{M}|=10k$(1K), $40k$(10K)]{
    \includegraphics[width=0.6\textwidth]{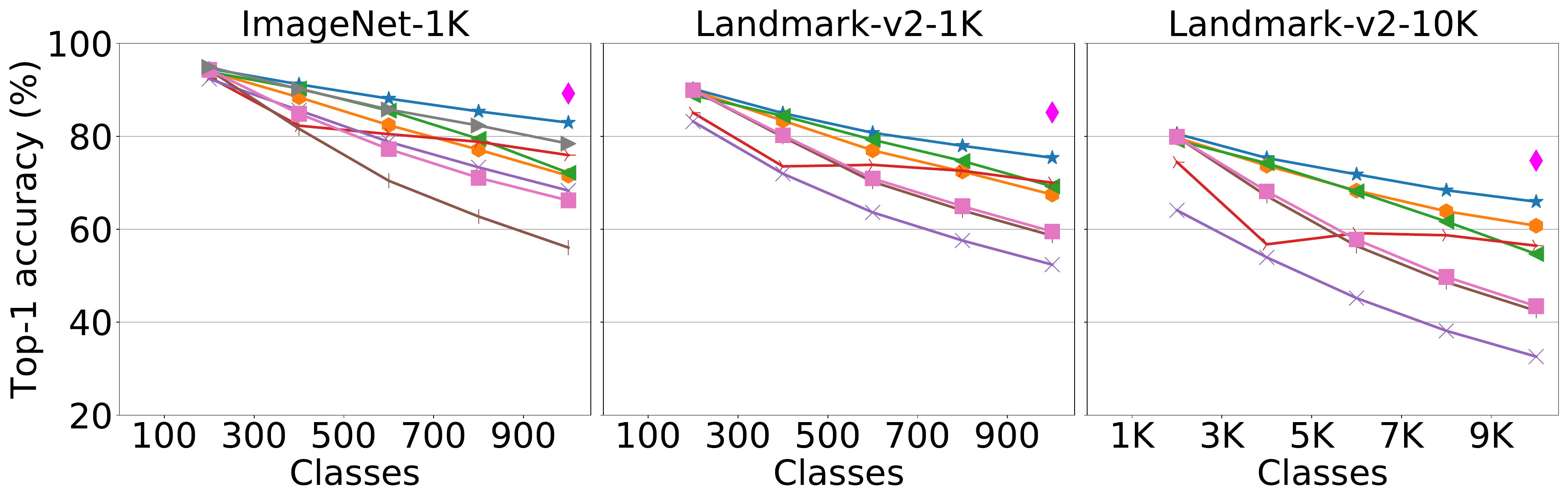}}
    \caption{Top-5 accuracy on ImageNet-1K, Landmark-1K and, Landmark-10K.}\label{fig:results_top_5}
    \vspace{-.8in}
\end{figure*}

We report overall Top-1 and Top-5 accuracy on each dataset with respect to the incremental task. By referring these figures below, we can compare each methods in a more class incremental view.
Figure \ref{fig:results_top_1} and \ref{fig:results_top_5} show the detailed results used to generate (Table 1 (Manuscript). Similarly, Figure \ref{fig:base and growing memory}, Figure \ref{fig:large base and fixed memory}, and Figure \ref{fig:large base and growing memory} show the overall results in Table \ref{table:growing_table}, Table \ref{table:largebase_table}, and Table \ref{table:both_table}. In summary, SS-IL achieves much higher accuracy than other baselines for most of the scenarios.

\begin{table*}[t]
\small
\centering
\vspace{-.0in}
\caption{The results on \textit{base} setting combined with \textit{growing memory} setting. $|\mathcal{M}_{per|}$ denotes the number of stored samples per old class. The evaluation metrics are the Average Top-1 and Top-5 accuracy. Accuracy is averaged over all the incremental tasks (i.e. including both initial task and incremental tasks)}\label{table:growing_table}
\resizebox{0.7\textwidth}{!}{\begin{tabular}{ccccc}
\hline
\multicolumn{1}{c|}{$ $}     & \multicolumn{2}{c|}{$T=10$}                                                                                                                            & \multicolumn{2}{c}{$|\mathcal{M}_{per}|=10$ (1K), $4$ (10K)}                                                                 \\ \hline\hline
\multicolumn{1}{c|}{Dataset} & \multicolumn{1}{c|}{ImageNet-1K}                 & \multicolumn{1}{c|}{Landmark-v2-1K}         & \multicolumn{1}{c|}{ImageNet-1K}       & Landmark-v2-1K      \\ \hline
\multicolumn{1}{c|}{$|\mathcal{M}_{per}|$}     & \multicolumn{1}{c|}{$5$ / $10$ / $20$}        & \multicolumn{1}{c|}{$5$ / $10$ / $20$}         & \multicolumn{1}{c|}{$T = 20$ / $T = 5$}   & \multicolumn{1}{c}{$T = 20$ / $T = 5$}  \\ \hline
                             & \multicolumn{4}{c}{Average Top-1 accuracy}                                                                                                                                                                                                                            \\ \hline\hline
\multicolumn{1}{c|}{EEIL \cite{(End2EndIL)Castro2018}}    & \multicolumn{1}{c|}{38.5 / 46.2 / 52.0}          & \multicolumn{1}{c|}{38.8 / 44.5 / 50.0}            & \multicolumn{1}{c|}{42.2 / 51.2}        & 38.8 / 49.5          \\
\multicolumn{1}{c|}{BiC \cite{(LargeScaleIL)Wu2019}}     & \multicolumn{1}{c|}{38.5 / 43.0 / 55.0}          & \multicolumn{1}{c|}{39.1 / 47.8 / 53.8}          & \multicolumn{1}{c|}{38.5 / 59.1}   & 36.4 / 58.1          \\
\multicolumn{1}{c|}{LUCIR \cite{(UnifiedIL)Hou2019}}     & \multicolumn{1}{c|}{47.0 / 49.7 / 52.7}          & \multicolumn{1}{c|}{46.1 / 49.3 / 52.6}   & \multicolumn{1}{c|}{39.0 / 59.5}   & 41.1 / 58.0          \\
\multicolumn{1}{c|}{PODNet \cite{(PODNet)Arthur}}     & \multicolumn{1}{c|}{44.2 / 57.3 / 56.7}          & \multicolumn{1}{c|}{-}      & \multicolumn{1}{c|}{40.5 / 63.6}     & -          \\
\multicolumn{1}{c|}{SS-IL (ours)}    & \multicolumn{1}{c|}{\textbf{62.3 / 63.4 / 63.9}} & \multicolumn{1}{c|}{\textbf{56.0 / 57.0 / 58.1} } & \multicolumn{1}{c|}{\textbf{57.0 / 67.4}} & \textbf{49.9 / 62.7} \\ \hline\hline
                             & \multicolumn{4}{c}{Average Top-5 accuracy}                                                                                                                                                                                                                            \\ \hline\hline
\multicolumn{1}{c|}{EEIL \cite{(End2EndIL)Castro2018}}    & \multicolumn{1}{c|}{65.3 / 72.3 / 76.9}          & \multicolumn{1}{c|}{58.5 / 64.3 / 69.5}               & \multicolumn{1}{c|}{68.1 / 76.0}    & 58.6 / 69.0          \\
\multicolumn{1}{c|}{BiC \cite{(LargeScaleIL)Wu2019}}     & \multicolumn{1}{c|}{57.5 / 67.1 / 78.2}          & \multicolumn{1}{c|}{57.0 / 67.4 / 72.8}          & \multicolumn{1}{c|}{59.8 / 79.8}           & 54.8 / 76.9         \\
\multicolumn{1}{c|}{LUCIR \cite{(UnifiedIL)Hou2019}}    & \multicolumn{1}{c|}{67.8 / 71.3 / 74.8}          & \multicolumn{1}{c|}{63.7 / 70.9 / 67.5}    & \multicolumn{1}{c|}{59.7 / 81.0}          & 59.2  /75.3          \\
\multicolumn{1}{c|}{PODNet \cite{(PODNet)Arthur}}   & \multicolumn{1}{c|}{64.8 / 79.2 / 78.9}          & \multicolumn{1}{c|}{-}        & \multicolumn{1}{c|}{62.7 / 84.2}        & -          \\
\multicolumn{1}{c|}{SS-IL (ours)}    & \multicolumn{1}{c|}{\textbf{85.3 / 85.9 / 86.0}} & \multicolumn{1}{c|}{\textbf{76.8 / 77.3 / 77.7}} & \multicolumn{1}{c|}{\textbf{81.6 / 88.2}} & \textbf{72.2 / 80.9} \\ \hline
\end{tabular}}
\vspace{-.2in}
\end{table*}

\begin{table*}[t]
\small
\centering
\vspace{-.0in}
\caption{The results on \textit{large base} setting combined with \textit{fixed memory} setting. $|\mathcal{M}|$ denotes the number of stored samples during training.}\vspace{-0.0in}\label{table:largebase_table}
\resizebox{0.7\textwidth}{!}{\begin{tabular}{ccccc}
\hline
\multicolumn{1}{c|}{$ $}     & \multicolumn{2}{c|}{$T=10$}                                                                                                                            & \multicolumn{2}{c}{$|\mathcal{M}|=10k$ (1K), $40k$ (10K)}                                                                 \\ \hline\hline
\multicolumn{1}{c|}{Dataset} & \multicolumn{1}{c|}{ImageNet-1K}                 & \multicolumn{1}{c|}{Landmark-v2-1K}       & \multicolumn{1}{c|}{ImageNet-1K}        & Landmark-v2-1K      \\ \hline
\multicolumn{1}{c|}{$|\mathcal{M}|$}     & \multicolumn{1}{c|}{$5k$ / $10k$ / $20k$}        & \multicolumn{1}{c|}{$5k$ / $10k$ / $20k$}           & \multicolumn{1}{c|}{$T = 20$ / $T = 5$}    & $T = 20$ / $T = 5$   \\ \hline
                             & \multicolumn{4}{c}{Average Top-1 accuracy}                                                                                                                                                                                                                            \\ \hline\hline
\multicolumn{1}{c|}{EEIL \cite{(End2EndIL)Castro2018}}    & \multicolumn{1}{c|}{40.6 / 46.2 / 50.7}          & \multicolumn{1}{c|}{41.0 / 46.6 / 51.6}       & \multicolumn{1}{c|}{39.3 / 52.7}           & 41.3 / 51.8         \\
\multicolumn{1}{c|}{BiC \cite{(LargeScaleIL)Wu2019}}     & \multicolumn{1}{c|}{41.4 / 46.4 / 50.7}          & \multicolumn{1}{c|}{41.1 / 45.8 / 49.8}        & \multicolumn{1}{c|}{36.2 / 55.4}         & 36.5 / 55.3         \\
\multicolumn{1}{c|}{LUCIR \cite{(UnifiedIL)Hou2019}}     & \multicolumn{1}{c|}{54.7 / 57.6 / 60.6}          & \multicolumn{1}{c|}{54.9 / 58.3 / 61.7}         & \multicolumn{1}{c|}{54.8 / 60.3}         & \textbf{55.2} / 61.4      \\
\multicolumn{1}{c|}{PODNet \cite{(PODNet)Arthur}}     & \multicolumn{1}{c|}{47.9 / 58.4 / \textbf{64.2}}          & \multicolumn{1}{c|}{-}        & \multicolumn{1}{c|}{51.0 / 65.3}                 & -          \\
\multicolumn{1}{c|}{SS-IL (ours)}    & \multicolumn{1}{c|}{\textbf{59.9 / 61.9} / 63.4}          & \multicolumn{1}{c|}{\textbf{57.5 / 59.8 / 61.9}}        & \multicolumn{1}{c|}{\textbf{57.1 / 65.7}}     & 55.0/ \textbf{64.0} \\ \hline\hline
                             & \multicolumn{4}{c}{Average Top-5 accuracy}                                                                                                                                                                                                                            \\ \hline\hline
\multicolumn{1}{c|}{EEIL \cite{(End2EndIL)Castro2018}}    & \multicolumn{1}{c|}{66.7 / 71.8 / 75.6}          & \multicolumn{1}{c|}{61.1 / 66.1 / 70.1}           & \multicolumn{1}{c|}{64.6 / 77.8}    & 60.6 / 71.1          \\
\multicolumn{1}{c|}{BiC \cite{(LargeScaleIL)Wu2019}}     & \multicolumn{1}{c|}{64.5 / 69.8 / 74.2}          & \multicolumn{1}{c|}{59.5 / 64.4 / 68.4}         & \multicolumn{1}{c|}{58.0 / 78.9}          & 53.6 / 74.0          \\
\multicolumn{1}{c|}{LUCIR \cite{(UnifiedIL)Hou2019}}    & \multicolumn{1}{c|}{77.5 / 80.5 / 83.2}          & \multicolumn{1}{c|}{73.3 / 76.3 / 78.9}   & \multicolumn{1}{c|}{78.4 / 82.6}    & 74.0 / 78.4         \\
\multicolumn{1}{c|}{PODNet \cite{(PODNet)Arthur}}   & \multicolumn{1}{c|}{68.9 / 79.9 / 84.7}          & \multicolumn{1}{c|}{-}                 & \multicolumn{1}{c|}{73.0 / 85.7}           & -          \\
\multicolumn{1}{c|}{SS-IL (ours)}    & \multicolumn{1}{c|}{\textbf{84.7 / 85.4 / 86.2}}          & \multicolumn{1}{c|}{\textbf{78.3 / 79.4 / 80.6}}       & \multicolumn{1}{c|}{\textbf{82.3 / 87.6}}     & \textbf{76.0 / 82.2} \\ \hline
\end{tabular}}
\vspace{-.2in}
\end{table*}

\begin{table*}[t]
\small
\centering
\vspace{-.0in}
\caption{The results on \textit{large base} setting combined with \textit{growing memory} setting. }\vspace{-0.0in}\label{table:both_table}
\resizebox{0.7\textwidth}{!}{\begin{tabular}{ccccc}
\hline
\multicolumn{1}{c|}{$ $}     & \multicolumn{2}{c|}{$T=10$}                                                                                                                            & \multicolumn{2}{c}{$|\mathcal{M}_{per}|=10$ (1K), $4$ (10K)}                                                                 \\ \hline\hline
\multicolumn{1}{c|}{Dataset} & \multicolumn{1}{c|}{ImageNet-1K}                 & \multicolumn{1}{c|}{Landmark-v2-1K}                & \multicolumn{1}{c|}{ImageNet-1K}          & Landmark-v2-1K      \\ \hline
\multicolumn{1}{c|}{$|\mathcal{M}_{per}|$}     & \multicolumn{1}{c|}{$5$ / $10$ / $20$}        & \multicolumn{1}{c|}{$5$ / $10$ / $20$}         & \multicolumn{1}{c|}{$T = 20$ / $T = 5$}   & $T = 20$ / $T = 5$   \\ \hline
                             & \multicolumn{4}{c}{Average Top-1 accuracy}                                                                                                                                                                                                                            \\ \hline\hline
\multicolumn{1}{c|}{EEIL \cite{(End2EndIL)Castro2018}}    & \multicolumn{1}{c|}{31.6 / 38.1 / 44.0}          & \multicolumn{1}{c|}{33.8 / 40.1 / 46.4}               & \multicolumn{1}{c|}{32.8 / 43.1}          & 36.6 / 44.7          \\
\multicolumn{1}{c|}{BiC \cite{(LargeScaleIL)Wu2019}}     & \multicolumn{1}{c|}{38.0 / 44.4 / -}          & \multicolumn{1}{c|}{39.4 / 44.6 / 49.0}              & \multicolumn{1}{c|}{33.9 / 53.6}    & 35.6 / 54.1          \\
\multicolumn{1}{c|}{LUCIR \cite{(UnifiedIL)Hou2019}}     & \multicolumn{1}{c|}{53.3 / 55.9 / 59.0}          & \multicolumn{1}{c|}{52.5 / 56.5 / 59.8}          & \multicolumn{1}{c|}{52.6 / 59.2}      &  \textbf{52.9} / 59.8          \\
\multicolumn{1}{c|}{PODNet \cite{(PODNet)Arthur}}     & \multicolumn{1}{c|}{41.9 / 53.5 / 61.9}          & \multicolumn{1}{c|}{-}             & \multicolumn{1}{c|}{45.7 / 63.3} & -          \\
\multicolumn{1}{c|}{SS-IL (ours)}    & \multicolumn{1}{c|}{\textbf{58.3 / 60.5 / 62.3}} & \multicolumn{1}{c|}{\textbf{55.7 / 58.0 / 60.2} } &  \multicolumn{1}{c|}{\textbf{55.4 / 64.9}} & \textbf{52.9} / \textbf{62.8} \\ \hline\hline
                             & \multicolumn{4}{c}{Average Top-5 accuracy}                                                                                                                                                                                                                            \\ \hline\hline
\multicolumn{1}{c|}{EEIL \cite{(End2EndIL)Castro2018}}    & \multicolumn{1}{c|}{56.9 / 64.1 / 69.6}          & \multicolumn{1}{c|}{52.6 / 59.1 / 65.0}       & \multicolumn{1}{c|}{57.3 / 69.0}        & 55.2 / 63.8          \\
\multicolumn{1}{c|}{BiC \cite{(LargeScaleIL)Wu2019}}     & \multicolumn{1}{c|}{61.2 / 68.0 / -}          & \multicolumn{1}{c|}{57.6 / 63.2 / 67.4}       & \multicolumn{1}{c|}{55.4 / 77.6}      & 52.6 / 73.0          \\
\multicolumn{1}{c|}{LUCIR \cite{(UnifiedIL)Hou2019}}    & \multicolumn{1}{c|}{75.7 / 78.8 / 81.8}          & \multicolumn{1}{c|}{70.1 / 75.0 / 77.6}      & \multicolumn{1}{c|}{76.4 / 81.5}             & 72.0 / 77.3          \\
\multicolumn{1}{c|}{PODNet \cite{(PODNet)Arthur}}   & \multicolumn{1}{c|}{62.0 / 75.3 / 83.0}          & \multicolumn{1}{c|}{-}       & \multicolumn{1}{c|}{67.6 / 84.4}              & -          \\
\multicolumn{1}{c|}{SS-IL (ours)}    & \multicolumn{1}{c|}{\textbf{84.1 / 84.8 / 85.7}} & \multicolumn{1}{c|}{\textbf{77.4 / 78.6 / 79.7}} & \multicolumn{1}{c|}{\textbf{81.2 / 87.5}} & \textbf{74.5 / 81.8} \\ \hline
\end{tabular}}
\vspace{-.2in}
\end{table*}

\begin{figure*}[t]
    \centering
    \subfigure[$T=10$ and $|\mathcal{M}_{per}|=5$]{
    \includegraphics[width=0.9\textwidth]{figures/base_memgrowing Top-1 & Top-5_5000_100.pdf}}
    \subfigure[$T=10$ and $|\mathcal{M}_{per}|=10$]{
    \includegraphics[width=0.9\textwidth]{figures/base_memgrowing Top-1 & Top-5_10000_100.pdf}}
    \subfigure[$T=10$ and $|\mathcal{M}_{per}|=20$]{
    \includegraphics[width=0.9\textwidth]{figures/base_memgrowing Top-1 & Top-5_20000_100.pdf}}
    \subfigure[$T=20$ and $|\mathcal{M}_{per}|=10$]{
    \includegraphics[width=0.9\textwidth]{figures/base_memgrowing Top-1 & Top-5_10000_50.pdf}}
    \subfigure[$T=5$ and $|\mathcal{M}_{per}|=10$]{
    \includegraphics[width=0.9\textwidth]{figures/base_memgrowing Top-1 & Top-5_10000_200.pdf}}
    \caption{Top-1 and Top-5 accuracy on base and growing memory setting at ImageNet-1K and Landmark-1K}\label{fig:base and growing memory}
\end{figure*}

\begin{figure*}[t]
    \centering
    \subfigure[$T=10$ and $|\mathcal{M}|=5$]{
    \includegraphics[width=0.9\textwidth]{figures/largebase_memfixed Top-1 & Top-5_5000_500_50.pdf}}
    \subfigure[$T=10$ and $|\mathcal{M}|=10k$]{
    \includegraphics[width=0.9\textwidth]{figures/largebase_memfixed Top-1 & Top-5_10000_500_50.pdf}}
    \subfigure[$T=10$ and $|\mathcal{M}|=20k$]{
    \includegraphics[width=0.9\textwidth]{figures/largebase_memfixed Top-1 & Top-5_20000_500_50.pdf}}
    \subfigure[$T=20$ and $|\mathcal{M}|=10k$]{
    \includegraphics[width=0.9\textwidth]{figures/largebase_memfixed Top-1 & Top-5_10000_500_25.pdf}}
    \subfigure[$T=5$ and $|\mathcal{M}|=10k$]{
    \includegraphics[width=0.9\textwidth]{figures/largebase_memfixed Top-1 & Top-5_10000_500_100.pdf}}
    \caption{Top-1 and Top-5 accuracy on large base and fixed memory setting at ImageNet-1K and Landmark-1K}\label{fig:large base and fixed memory}
\end{figure*}

\begin{figure*}[t]
    \centering
    \subfigure[$T=10$ and $|\mathcal{M}_{per}|=5k$]{
    \includegraphics[width=0.9\textwidth]{figures/largebase_memgrowing Top-1 & Top-5_5000_500_50.pdf}}
    \subfigure[$T=10$ and $|\mathcal{M}_{per}|=10$]{
    \includegraphics[width=0.9\textwidth]{figures/largebase_memgrowing Top-1 & Top-5_10000_500_50.pdf}}
    \subfigure[$T=10$ and $|\mathcal{M}_{per}|=20$]{
    \includegraphics[width=0.9\textwidth]{figures/largebase_memgrowing Top-1 & Top-5_20000_500_50.pdf}}
    \subfigure[$T=20$ and $|\mathcal{M}_{per}|=10$]{
    \includegraphics[width=0.9\textwidth]{figures/largebase_memgrowing Top-1 & Top-5_10000_500_25.pdf}}
    \subfigure[$T=5$ and $|\mathcal{M}_{per}|=10$]{
    \includegraphics[width=0.9\textwidth]{figures/largebase_memgrowing Top-1 & Top-5_10000_500_100.pdf}}
    \caption{Top-1 and Top-5 accuracy on large base and growing memory setting at ImageNet-1K and Landmark-1K}\label{fig:large base and growing memory}
\end{figure*}

\clearpage
{\small
\bibliographystyle{ieee_fullname}
\bibliography{bibfile}
}